%% file: acl_main.tex
\pdfoutput=1

\documentclass[11pt]{article}
\usepackage{tcolorbox}

\usepackage[final]{acl}

\usepackage{times}
\usepackage{latexsym}

\usepackage[T1]{fontenc}

\usepackage[utf8]{inputenc}

\usepackage{microtype}

\usepackage{inconsolata}

\usepackage{graphicx}

\usepackage{booktabs} 
\usepackage{multirow}
\usepackage{listings}
\usepackage{hyperref}
\usepackage{amsmath}

\newcommand{\ours}{\texorpdfstring{\includegraphics[height=1em]{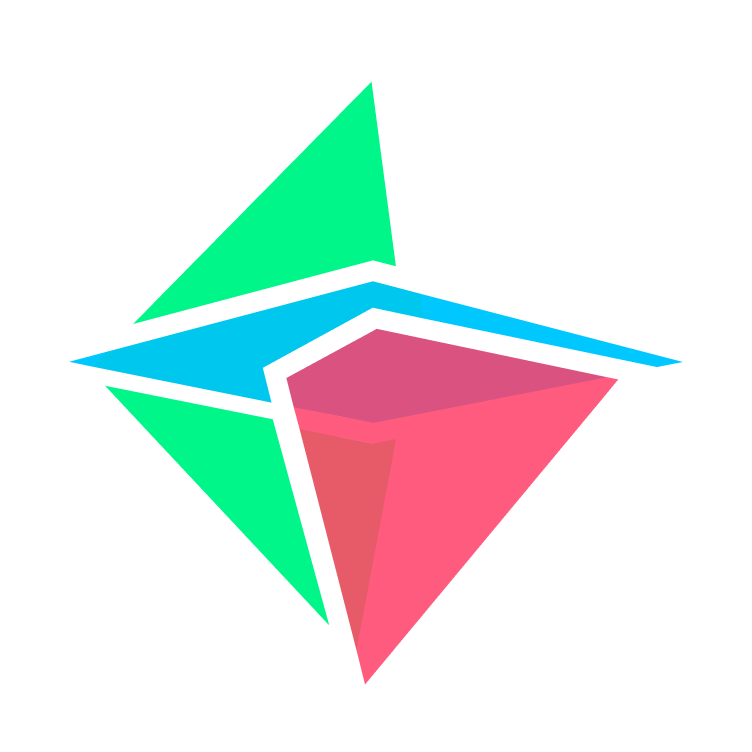}\textbf{FADE}}{FADE}}
\newcommand{\repolink}{\url{https://github.com/brunibrun/FADE}}

\title{\ours: Why Bad Descriptions Happen to Good Features}

\author{
 \textbf{Bruno Puri\textsuperscript{1,2,*}},
 \textbf{Aakriti Jain\textsuperscript{1,*}},
 \textbf{Elena Golimblevskaia\textsuperscript{1,*}},
 \textbf{Patrick Kahardipraja\textsuperscript{1}},
 \\
 \textbf{Thomas Wiegand\textsuperscript{1,2,3}},
 \textbf{Wojciech Samek\textsuperscript{1,2,3}},
 \textbf{Sebastian Lapuschkin\textsuperscript{1,4}}
\\
\\
\textsuperscript{1}Department of Artificial Intelligence, Fraunhofer Heinrich Hertz Institute\\
\textsuperscript{2}Department of Electrical Engineering and Computer Science, Technische Universität Berlin\\
\textsuperscript{3}BIFOLD - Berlin Institute for the Foundations of Learning and Data\\
\textsuperscript{4}Centre of eXplainable Artificial Intelligence, Technological University Dublin
\\
 \small{
   \textbf{corresponding authors:} \texttt{\{wojciech.samek,sebastian.lapuschkin\}@hhi.fraunhofer.de}}
}

\begin{document}
\maketitle

\begingroup
\renewcommand\thefootnote{*}
\footnotetext{These authors contributed equally.}
\endgroup

\input{sections/Abstract}
\input{sections/1_Introduction}
\input{sections/2_Related_Work}
\input{sections/3_Evaluation}
\input{sections/4_Experiments}
\input{sections/5_Conclusion}
\input{sections/Limitations}
\input{sections/Acknowledgements}

\bibliography{bibliography}
\clearpage
\newpage

\appendix

\input{sections/Appendix}

\end{document}

%% file: sections/Abstract.tex
\begin{abstract}
Recent advances in mechanistic interpretability have highlighted the potential of automating interpretability pipelines in analyzing the latent representations within LLMs. While this may enhance our understanding of internal mechanisms, the field lacks standardized evaluation methods for assessing the validity of discovered features. We attempt to bridge this gap by introducing \textbf{\ours: Feature Alignment to Description Evaluation}, a scalable model-agnostic framework for automatically evaluating feature-to-description alignment. FADE evaluates alignment across four key metrics -- \textit{Clarity, Responsiveness, Purity, and Faithfulness} -- and systematically quantifies the causes of the misalignment between features and their descriptions. We apply \ours \ to analyze existing open-source feature descriptions and assess key components of automated interpretability pipelines, aiming to enhance the quality of descriptions. Our findings highlight fundamental challenges in generating feature descriptions, particularly for SAEs compared to MLP neurons, providing insights into the limitations and future directions of automated interpretability. We release \ours\ as an open-source package at: \url{https://github.com/brunibrun/FADE}. 
\end{abstract}

%% file: sections/1_Introduction.tex
\section{Introduction}

Understanding the latent features of machine learning models and aligning their descriptions with human-comprehensible concepts remains a crucial challenge in AI interpretability research. Recent advances have made significant strides in this direction, by introducing automated interpretability methods \cite{bills2023language, bykov2024labeling, choi2024automatic} that leverage larger language-capable models to describe the latent representations of smaller models \cite{DBLP:journals/tmlr/BykovDGMH23, templeton2024scaling, dreyer2025mechanistic}. This facilitates inspection of ML models, enabling a deeper understanding of models' behavior which enhances our ability to identify or mitigate harmful responses and biases \cite{lee2024mechanistic, gandikota2024erasing}.
\begin{figure*}[t!]
    \centering
    \includegraphics[width=1\linewidth]{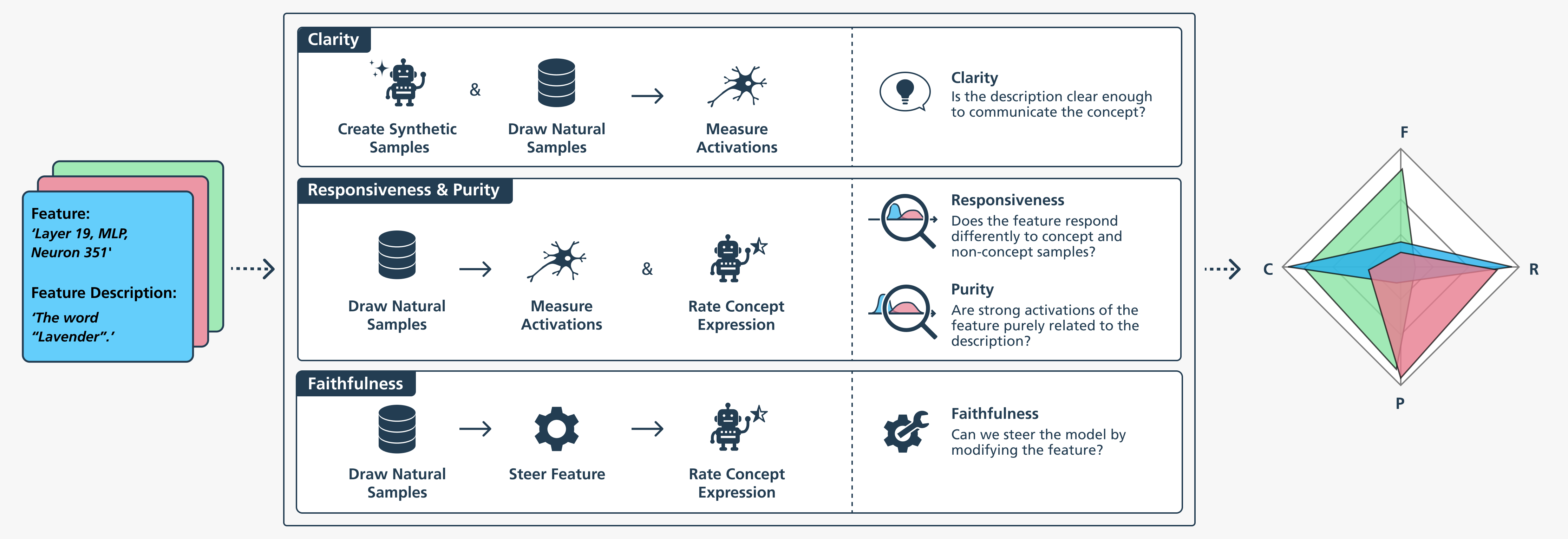} 
    \caption{Visualization of the \ours \ pipeline for three features and their corresponding feature descriptions.
    }
    \label{fig:evaluation-pipeline-graphic}
\end{figure*}
A key insight from these investigations is the highly polysemantic nature of individual neurons -- they rarely correspond to single, clear concepts. This discovery has led to the development and adoption of sparse autoencoders (SAEs) \cite{ goodfire, bricken2023monosemanticity, rajamanoharan2024jumpingaheadimprovingreconstruction}, which are intended to decompose polysemantic representations by separating neuron activations into more interpretable components. While SAEs offer a promising approach for feature decomposition, their reliability remains an open question. Recent research reveals significant variability in the way SAEs capture the underlying learned features \cite{heap2025sparse, paulo2025sparse}, thus highlighting the need for a holistic framework for the evaluation of feature-to-description alignment.
 To the best of our knowledge, there is an absence of widely accepted quantitative metrics for evaluating the quality and effectiveness of open-vocabulary feature descriptions.
 Different methodologies rely on custom evaluation criteria which makes it challenging to conduct meaningful, generalizable comparisons across techniques. Additionally, existing evaluation approaches typically optimize for a single metric \cite{bills2023language, choi2024automatic} which may not capture the full complexity of a feature's behavior and leaves open questions about whether the model truly encodes the hypothesized concept rather than simply correlating with the measured feature. With our work, we contribute as follows:\newline

\textbf{[1]} We present a robust automated evaluation framework designed for broad applicability across model architectures and their SAE implementations.\ours \ combines four metrics that allow quantitative analysis of different aspects of alignment between features and their generated descriptions.\footnote{We release \ours \ as an open-source Python package available at \repolink, and include example notebooks as well as some of the feature evaluations presented in this work.} \newline

\textbf{[2]} Through systematic empirical analysis, we provide insights into how various components of the autointerpretability pipeline -- such as the number of layers, sample sizes, and architectural choices -- affect the quality of feature descriptions.\newline

\textbf{[3]} We release a set of feature descriptions, presented as part of this work for \texttt{Gemma-2-2b} and \texttt{Gemma Scope} SAEs along with their evaluations.

%% file: sections/2_Related_Work.tex
\section{Related Work}

Evaluating the alignment between features and their descriptions has become increasingly important with the rise of automated interpretability approaches. While manually inspecting highly activating examples remains a common method to validate interpretations and demonstrate automated interpretability techniques \cite{bills2023language, templeton2024scaling}, more scalable tools are needed for thorough quantitative evaluation. Many automated or semi-automated approaches have been proposed, generally falling into activation-centric and output-centric methods.

Activation-centric methods focus on measuring how well a feature’s activations correspond to its assigned description.
One prominent approach is a simulation-based scoring, where an LLM predicts feature activations based on the description and input data, and the correlation between predicted and real activations of a feature is measured \cite{bills2023language, bricken2023monosemanticity, choi2024automatic}. While elegant, this approach can be computationally expensive and tends to favor broad, high-level explanations.
A related and conceptually more straightforward way to measure how well the description explains a feature's behavior is to try to directly generate synthetic samples using the description and compare the resulting activations between concept and non-concept samples \cite{huang-etal-2023-rigorously, DBLP:conf/nips/KopfBHLHB24, gurarieh2025enhancingautomatedinterpretabilityoutputcentric, shaham2025multimodalautomatedinterpretabilityagent}. However, generated datasets are typically small (on the order of 5–20 samples \cite{huang-etal-2023-rigorously, gurarieh2025enhancingautomatedinterpretabilityoutputcentric}) and often constrained to rigid syntactic structures or focus only on the occurrence of particular tokens, making them less effective for evaluating abstract or open-ended language concepts \cite{huang-etal-2023-rigorously, foote2023n2gscalableapproachquantifying}.
Another strategy is rating individual samples from a natural dataset for how strongly they express a concept and compare those ratings to the feature's activations \cite{huang-etal-2023-rigorously, paulo2024automaticallyinterpretingmillionsfeatures, templeton2024scaling}. 
A common limitation of activation-centric methods is that they primarily assume activations to be positively correlated with the concept, thus ignoring negatively encoded features \cite{huang-etal-2023-rigorously, DBLP:conf/nips/KopfBHLHB24}.

Output-centric methods instead assess how feature activations influence model behavior. Some approaches measure the general decrease in performance of the model after ablating the feature \cite{bills2023language, makelov2024principledevaluationssparseautoencoders}, while others use steering-based interventions, where an increase in generated outputs containing the concept is used as a proxy for feature alignment \cite{paulo2024automaticallyinterpretingmillionsfeatures, gurarieh2025enhancingautomatedinterpretabilityoutputcentric}.

There is a growing need for frameworks that integrate multiple perspectives to provide a comprehensive assessment of feature-to-description alignment across different interpretability methods.
For instance, prior work \cite{bills2023language, menon2025analyzinginabilitiessaesformal, gurarieh2025enhancingautomatedinterpretabilityoutputcentric} has shown that while activation-centric and output-centric measures often correlate, they do not necessarily imply a causal relationship.
Some studies focus exclusively on SAEs \cite{goodfire, paulo2024automaticallyinterpretingmillionsfeatures}, while others analyze MLP neurons \cite{bills2023language, choi2024automatic}. Developing an architecture-agnostic framework for feature-description evaluation is essential for enabling robust quantitative comparisons across interpretability approaches.
Although efforts have been made to integrate multiple evaluation perspectives \cite{paulo2024automaticallyinterpretingmillionsfeatures, gurarieh2025enhancingautomatedinterpretabilityoutputcentric}, these remain fragmented and are often too narrowly scoped to handle open-ended language descriptions.

Our work addresses these limitations by introducing a more comprehensive evaluation framework that combines activation- and output-centric metrics while explicitly considering interpretability for open-ended language descriptions.

%% file: sections/3_Evaluation.tex
\section{Evaluating Feature Explanations} \label{sec:methodology}

\begin{figure*}[ht]
    \centering
    \includegraphics[width=1\linewidth]{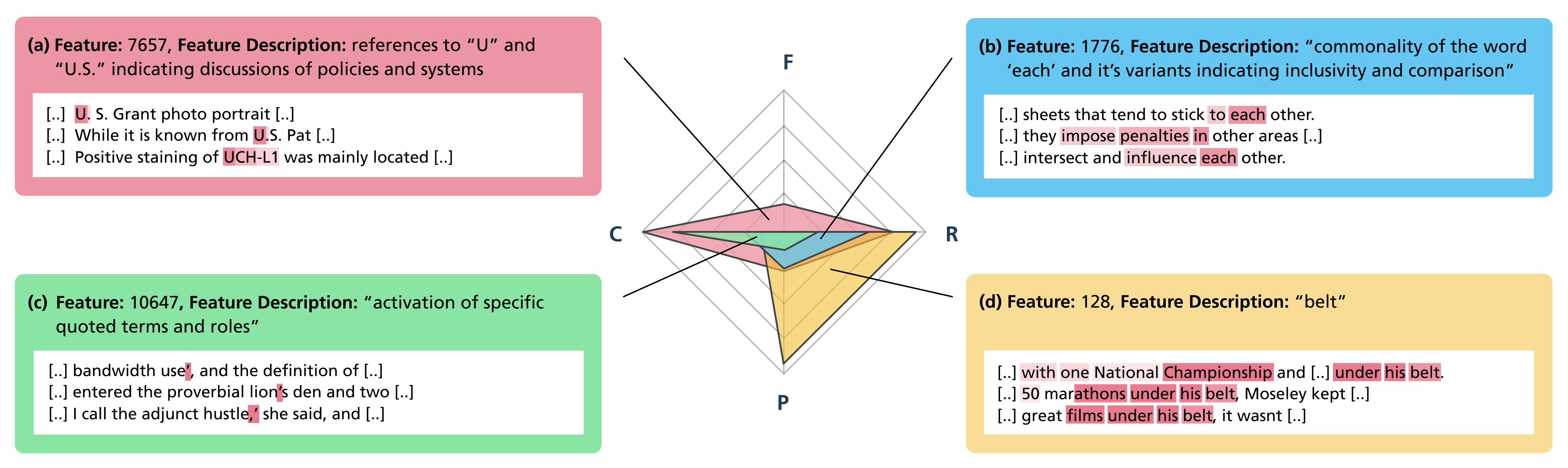}
    \caption{ \ours \ can highlight different problems that arise with description generation. (a) The description for feature 7657 catches its main concept, low Purity indicates that this feature is clearly polysemantic.
(b) Feature 1776 strongly reacts to the word ``each'', but also to many other general, unrelated words, resulting in lower Clarity and Purity. (c) Description for feature 10647 is expressed too broad, resulting in low Responsiveness and Purity.  (d) Feature 128 encodes the concept ``under someone's belt''. However, the derived description ``belt'', is not clear or specific enough to be useful for generating synthetic data that would activate the feature, but it is still closely related to the concept, therefore Responsiveness and Purity are high. 
}
    \label{fig:qualitative-examples}
\end{figure*}

Our primary objective is to establish a comprehensive framework that automatically evaluates feature descriptions across a variety of feature types without human intervention. Our framework encompasses four distinct metrics: \textbf{Clarity}, \textbf{Responsiveness}, \textbf{Purity}, and \textbf{Faithfulness}, which we consider necessary and sufficient for assessing the alignment between a feature and its description. In our opinion, such a comprehensive evaluation framework is necessary to ensure that features encode the ascribed concept in a robust way. As feature descriptions are often generated by optimizing for a single metric, such as maximizing the activations of specific neurons, they do not necessarily generalize well to other quantifiable aspects, such as Faithfulness \cite{bills2023language, choi2024automatic}.

We base our approach on four key assumptions. First, we adopt a \raisebox{.5pt}{\textcircled{\footnotesize 1}} \textit{Binary Concept Expression} model, whereby a concept is either present in a text sequence or absent. Second, we assume \raisebox{.5pt}{\textcircled{\footnotesize 2}} \textit{Concept Sparsity}, i.e.\ that a given concept appears only rarely in natural datasets, though a sufficiently large dataset will contain some representative examples.
Third, we assume \raisebox{.5pt}{\textcircled{\footnotesize 3}} \textit{Feature Reactivity}, meaning, when a feature encodes a concept, its activations are significantly stronger on samples that express the concept.
This will be valid especially for SAEs, since by construction, for most samples, their activations are zero.
This is a strong assumption, as it also implies that a feature should activate strongly only for a single concept.
Note, however, that this does not require strict monosemanticity \cite{bricken2023monosemanticity}.
In our framework a feature might encode multiple, even entirely unrelated topics, as long as its feature description fully describes all of them. Unlike traditional monosemanticity, which assumes features should directly align with a single human-interpretable category, our framework evaluates interpretability based on whether the feature description accurately reflects the feature’s truly encoded concept, rather than enforcing human-aligned conceptual boundaries.
Assumption \raisebox{.5pt}{\textcircled{\footnotesize 1}} and \raisebox{.5pt}{\textcircled{\footnotesize 3}} allow us to interpret the activations of a feature as output of a ``classifier'' of the encoded concept, which can then be easily evaluated. For our metrics, we expect a feature to encode the concept linearly in its activations. Finally, we assume \raisebox{.5pt}{\textcircled{\footnotesize 4}} \textit{Causality} -- a feature is expected to causally influence the model’s output so that modifying its activation will lead to predictable changes in the generation of concept-related content. These four assumptions will not always hold but are necessary simplifications for now.

\subsection{Evaluation Framework Components}

Our evaluation framework consists of three main components: A \textit{subject LLM}, that contains the features we want to evaluate, a \textit{natural dataset}, that ideally should be close to the LLM training data distribution and is sufficiently large to contain all the concepts, of which the descriptions we want to evaluate, and an \textit{evaluating LLM}, an open- or closed-source LLM that is used for automating the evaluation process. The evaluating LLM is used for ``human-like'' inference tasks, such as rating the strength of concept expression in samples and creating synthetic concept data.

\subsection{Evaluation Metrics}

\paragraph{Clarity} evaluates whether a feature’s description is precise enough to generate strongly activating samples. We assess this by prompting the evaluating LLM to generate synthetic samples based on the feature description (see prompts in Appendix~\ref{appendix-evaluation-experiment-setup-prompts}). Unlike \citet{gurarieh2025enhancingautomatedinterpretabilityoutputcentric}, which generates non-concept sequences artificially, we sample them uniformly from the natural dataset to avoid unnatural biases (i.e. by asking the evaluating LLM not to think about pink elephants). If a feature is well explained by its description, the synthetic concept samples should elicit significantly stronger activations than non-concept samples. We quantify this separability using the absolute Gini coefficient

{\scriptsize
    \begin{align}
    \mathrm{G_{abs}}(A_{c}, A_{n}) = \left| 2 \cdot \left( \frac{\sum\limits_{a_c \in A_c} \sum\limits_{a_n \in A_n} \mathbf{1}_{[a_c > a_n]}}{\| A_c \|_0 \cdot \| A_n \|_0} \right) -1 \right|
   \end{align}
}
\\
where $A_c$ and $A_n$ are the sets of concept and non-concept activations, respectively. Since this metric focuses on linear separability rather than precision, it remains robust even when concept samples occasionally appear within the natural dataset. A low Clarity score indicates that either the description is not precise enough to be useful, or might simply be unfitting for the feature, resulting in similar activations for both concept and non-concept samples. For example, in Figure~\ref{fig:qualitative-examples}, feature (d) responds to ``having something under one’s belt,'' yet is inaccurately described as ``belt''. Conversely, a high Clarity score confirms that we can effectively generate samples that elicit strong activations in the feature, although it does not guarantee that the feature is monosemantic or causally involved.

\paragraph{Responsiveness} evaluates the difference in activations between concept and non-concept samples. We select samples from the natural dataset based on their activation levels, drawing both from the highest activations and from lower percentiles (details in Appendix~\ref{appendix-evaluation-implementation}). Following an approach similar to \citet{templeton2024scaling}, we prompt the evaluating LLM to rate each sample on a three-point scale to indicate how strongly the concept is present (0 = not expressed, 1 = partially expressed, 2 = clearly expressed). By discarding the ambiguous (partially expressed) cases, we effectively binarize samples into concept and non-concept categories. We compute the Responsiveness score again using the absolute Gini coefficient. A low Responsiveness score indicates that activations of concept-samples are similarly strong as non-concept samples, while a high score indicates that, in natural data, samples with strong activations reliably contain the concept.

\paragraph{Purity} is computed using the same set of rated natural samples as Responsiveness, but with a different focus: it evaluates whether the strong activations are exclusive to the target concept. In contrast to \cite{huang-etal-2023-rigorously}, who measure recall and precision for a single threshold, we measure the Purity using the Average Precision (AP)

{\scriptsize
    \begin{align}
    \mathrm{AP}(A_{c}, A_{n}) = \sum_{j} \left( r_{j} - r_{j-1} \right) \cdot p_{j}
    \end{align}
}
\\
where $r_j$ is the recall and $p_j$ is the precision computed at threshold $j$, for each possible threshold, based on $A_c$ and $A_n$. The AP penalizes instances where non-concept samples also trigger high activations. A Purity score near one thus indicates that the feature’s activations are highly specific to the concept, whereas a score near zero suggests that top activations occur for other unrelated concepts as well. This is, for example, the case in polysemanticity, where a feature responds to multiple unrelated concepts. 

\paragraph{Faithfulness} addresses the causal relationship between a feature and the model’s output. In other words, it tests whether direct manipulation of the feature’s activations can steer the model’s output toward generating more concept-related content. To evaluate Faithfulness, we take random samples from the natural dataset and have the subject LLM generate continuations while applying different modifications to the feature’s activation.
For neurons, we multiply the raw activations by a range of factors, including negative values, so that we do not impose a directional bias on how the concept is encoded. For SAE features, of which the activations are more sparse, we first determine the maximum activation observed in the natural dataset \cite{templeton2024scaling} and then scale this value by the different modification factors. After generating the modified continuations, the evaluating LLM rates how strongly the concept appears in each output.
We quantify the strength of this causal influence by measuring the largest increase we were able to steer the model in producing concept-related outputs

{\scriptsize
    \begin{align}
    \mathrm{Faithfulness}(\mathbf{R}) = \dfrac{\max(\max(\mathbf{R}) - R_0, 0)}{1-R_0}
    \end{align} 
}
\\
where $\mathbf{R}$ is a vector capturing the proportion of concept-related outputs for each modification factor, and $R_0$ denotes the base case in which the feature is “zeroed out” (i.e., multiplied by zero). A Faithfulness score of zero implies that manipulating the feature does not increase the occurrence of concept-related outputs, while a score of one indicates that for some modification factor the concept is produced in every continuation.

\subsection{\ours\ Evaluation Framework}

\ours\ is designed to work with a wide variety of Subject Models, including most HuggingFace Transformers \citep{wolf-etal-2020-transformers} as well as any feature implemented as a named module in PyTorch. Moreover, it is extensible: interpretability tools such as SAEs or various supervised interpretability techniques can be integrated with minimal effort, provided they implement some basic steering functionality. In addition, we offer an interface for a diverse set of evaluating LLMs, whether open-weight or proprietary, with support for frameworks such as vLLM, Ollama, OpenAI, and Azure APIs.

%% file: sections/4_Experiments.tex
\section{Experiments}

In this section, we apply \ours\ to assess the alignment of features and their descriptions generated via various state-of-the-art automated interpretability methods \cite{choi2024automatic, lieberum-etal-2024-gemma}. Our goal is to demonstrate that the proposed framework provides a robust, multidimensional measure of feature-to-description alignment.

\paragraph{Experimental Setup}
As a natural dataset for the evaluations we use samples drawn from the test partition of the Pile dataset \cite{gao2020pile}, preprocessed as shown in 
Appendix~\ref{appendix-dataset}, giving us approximately 5 million samples.
As evaluating LLM we use the OpenAI model \texttt{gpt-4o-mini-2024-07-18} unless stated otherwise. Prompts for the evaluating LLM as well as details on the hyperparameters can be found in Appendix~\ref{appendix-evaluation-experiment-setup}. We run the experiments on $10^3$ randomly chosen features from a single layer of a model: layer 20 for \texttt{Gemma-2-2b} \cite{gemmateam2024gemma2improvingopen}, layer 20 of \texttt{Gemma Scope} SAEs \cite{lieberum-etal-2024-gemma}, and layer 19 of \texttt{Llama-3.1-8B-Instruct} \cite{grattafiori2024llama3herdmodels} (see Appendix~\ref{appendix-autointerpretability-pipeline} for details). 
The evaluation results have a high variance, which is caused by both the inherent difficulty of interpreting some features as well as the quality of the ascribed feature descriptions. 
Therefore the mean values are reported only where they help interpret the distribution of the metrics. For all of the presented tables we demonstrate the full distributions as kernel-density estimations with bandwidth adjustment factor 0.3 in Appendix~\ref{appendix-extended-results}. 

\paragraph{Automated interpretability approach}
Feature descriptions, which we refer to as MaxAct*, are generated based on samples of the train partition of the Pile dataset, that demonstrate maximum activation on the feature, similarly to methods utilized in \cite{bills2023language, paulo2024automaticallyinterpretingmillionsfeatures, rajamanoharan2024jumpingaheadimprovingreconstruction}. The minor differences between MaxAct and MaxAct* are prompts, optimized on qualitative analysis provided via \ours, and preprocessing steps of the dataset. The automated interpretability pipeline is described in Appendix~\ref{appendix-autointerpretability-pipeline}.

\subsection{Depth and Reliability of Evaluations}\label{sec:Results}

\begin{figure*}[t!]
    \centering
    \includegraphics[width=1\linewidth]{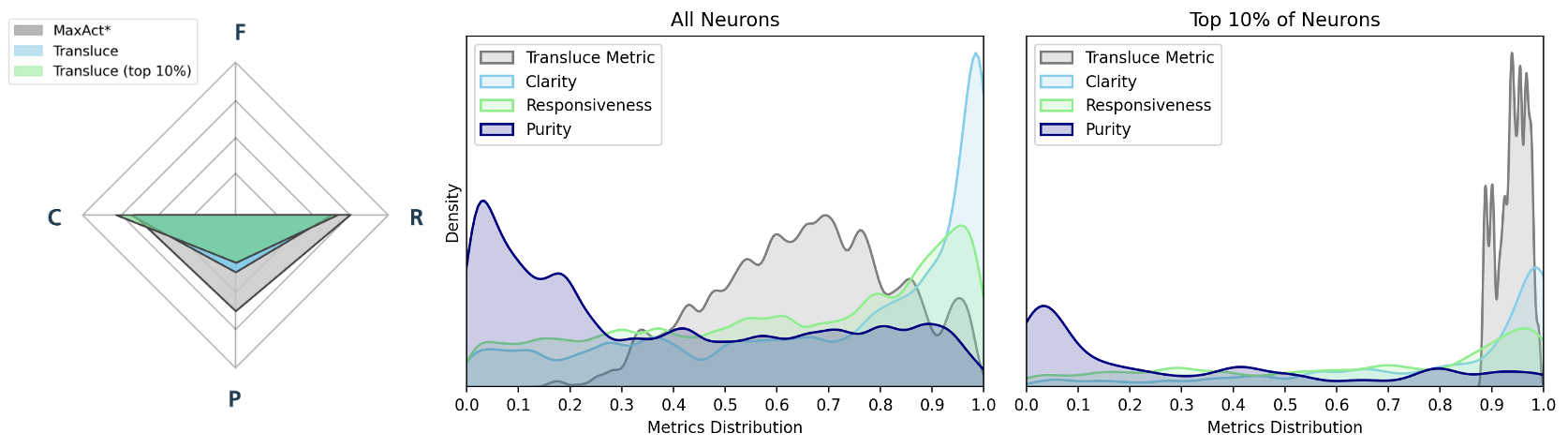}
    \caption{Distribution of metrics in \ours\ framework and a simulation-based metric on a uniformly subsampled set of features and the top 10\% of subsampled features, selected based on the simulated activation metric generated in \cite{choi2024automatic} (right), and comparison to the proposed descriptions for these features, generated in this work (left).}
    \label{fig:metrics-comparison}
\end{figure*}

\paragraph{Limitations of single-metric approaches}
We compare \ours\ with simulated-activation-based metrics \cite{bills2023language, templeton2024scaling, choi2024automatic}, that, while computationally efficient, 
fail to fully capture the feature-to-description alignment, potentially overlooking critical issues like polysemanticity.
To illustrate this, we analyze feature descriptions of \texttt{Llama-3.1-8B-Instruct} generated in \cite{choi2024automatic}. As shown in Figure~\ref{fig:metrics-comparison}, despite high simulated-activation scores, \ours\ identifies many features with low Purity. Moreover, comparing the average results across all subsampled features with the top 10\% of features based on the simulated-activation metric, we find only a marginal gain in Clarity and Responsiveness, while the Purity worsens.
Via MaxAct*, we generate descriptions with slightly lower Clarity but significantly higher Responsiveness and Purity. We attribute this to the Explainer Model in \cite{choi2024automatic} being fine-tuned on descriptions optimized for the simulated-activation metric, which aligns more closely with Clarity but neglects Responsiveness and Purity.

\paragraph{Better models provide better evaluations}
As the evaluating LLM is one of the most computationally expensive components of our framework, selecting a model that balances performance and cost is critical. Larger models generally achieve better performance, but at significantly higher computational costs. To determine a minimal feasible model size and capability required for effective evaluation, we conduct a quantitative analysis of concept-expression ratings across various open-weight and proprietary models, using \texttt{GPT-4o} as a baseline due to its superior benchmark performance \cite{openai2024gpt4o}. 
We evaluate models using Neuronpedia \cite{neuronpedia} feature descriptions for the \texttt{Gemma Scope} SAEs, generated via the MaxAct method \cite{rajamanoharan2024jumpingaheadimprovingreconstruction}. By comparing deviations in concept strength ratings between \texttt{GPT-4o} and other models, we assess their relative performance (see Table~\ref{tab:evaluation-models-comparison}).

\begin{table}
    \scriptsize
    \centering
    \begin{tabular}{lcccc}
        Model & Class 0 & Class 1 & Class 2 & Valid \\
        \hline
        \hline
        \noalign{\vskip .5mm}  
        GPT-4o & 243,233 & 24,766 & 19,716 & 100 \\
        \hline
                \noalign{\vskip .5mm}  
        Llama-3.2-1B & 63.8 & 22.1 & 20.9 & 8.8\\
        Llama-3.2-3B & 77.4 & 9.9 & 70.3 & 72.2 \\
        Llama-3.1-8B & 82.0 & 14.6 & 85.6 & 82.8\\
        Llama-3.3-70B 4q & 88.7 & 31.5 & \textbf{92.6} & 88.3\\
        GPT-4o mini & \textbf{93.4} & \textbf{44.8} & 79.3 & \textbf{88.6} \\
        \hline
    \end{tabular}
    \caption{Concept rating procedure for different evaluating LLMs. The \texttt{GPT-4o} baseline shows the number of occurrences per class. The other models show their alignment with the \texttt{GPT-4o} rating in percent. The ``Valid'' column shows the percentage of samples that were correctly classified. Class 0 represents no alignment with the concept, class 1 a partial alignment and class 2 means the samples clearly exhibit the concept.}
    \label{tab:evaluation-models-comparison}
\end{table}
\begin{figure*}[h]
    \centering
    \includegraphics[width=1\linewidth]{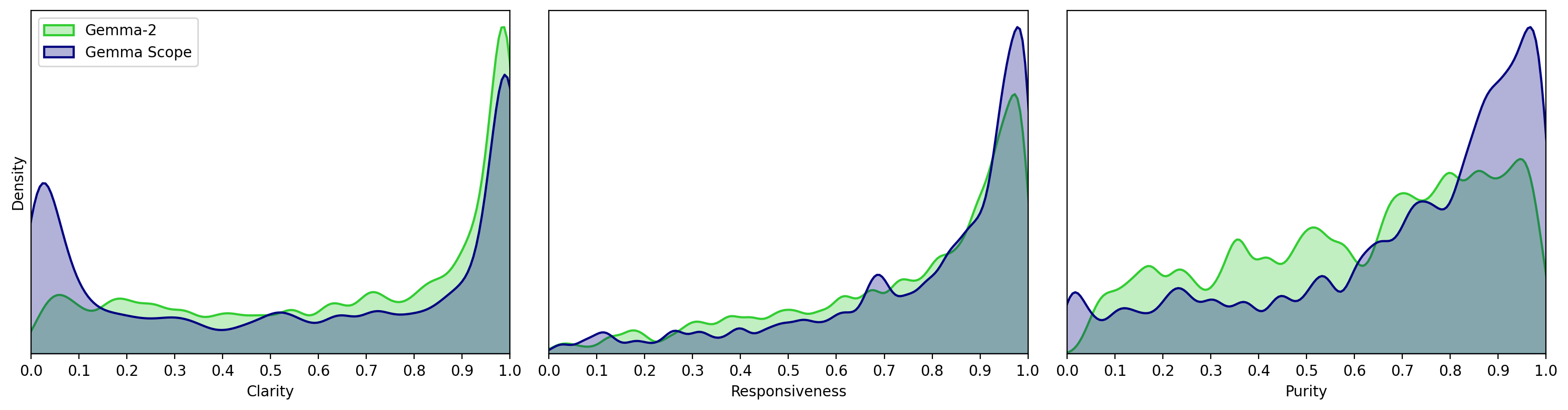}
    \caption{Feature description fit for neuron-based features (\texttt{Gemma-2}) and SAE-based features (\texttt{Gemma Scope}).}
    \label{fig:neurons-vs-saes}
\end{figure*}

\begin{figure*}
    \centering
    \includegraphics[width=1\linewidth]{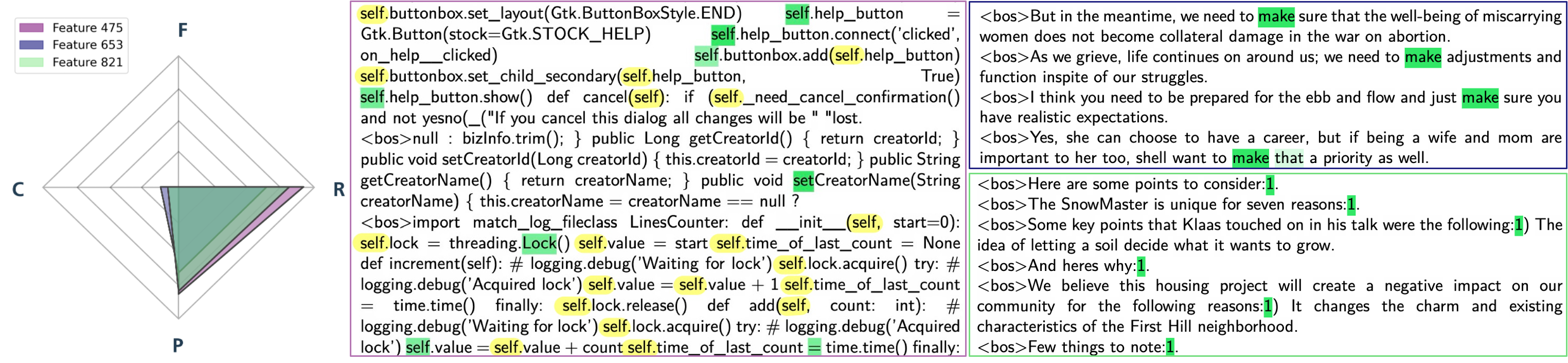}
    \caption{Examples of feature descriptions, obtained via MaxAct*, demonstrating low Clarity, but medium or high Responsiveness and Purity. Tokens are highlighted in green when the feature is activated, and in yellow if the feature is not activated even though the concept is present. Feature 475: ``Frequent presence of token 'self' indicating object oriented programming concept''. Feature 653: ``Word 'make' in different forms, expressions and concepts''. Feature 821: ``Expressions indicating lists or explanations''. }
    \label{fig:low-calrity-heatmaps}
\end{figure*}
Our findings reveal a clear trend: larger, more capable models consistently yield better evaluations. The open-weight \texttt{Llama-3.3-70B-Instruct} (AWQ 4-bit quantized) performs comparably to the proprietary \texttt{GPT-4o mini}, a widely used model in autointerpretability research \cite{choi2024automatic, neuronpedia}. While Class 1 (partial alignment) is the most error-prone, smaller models, such as \texttt{Llama-3.2-3B-Instruct}, remain viable for the more critical Class 0 (no alignment) and Class 2 (strong alignment). However, for optimal performance, models smaller than \texttt{Llama-3.1-8B-Instruct} are likely insufficient.

\paragraph{Generating feature descriptions for SAEs is more challenging than for MLP neurons} 
To compare MLP neurons and SAE features, we analyze \texttt{Gemma-2-2b} and \texttt{Gemma Scope} SAEs. While \texttt{Gemma-2} outperforms \texttt{Gemma Scope} SAEs in average Clarity (see Table~\ref{tab:numberical-vs-delimeters}), the Clarity score distribution reveals a left-skewed peak for SAEs, as depicted in Figure~\ref{fig:neurons-vs-saes}. Further analysis identifies a cluster of features with low Clarity but moderate to high Responsiveness and Purity. These features have descriptions that approximate the encoded concept but lack the precision to strongly activate the SAE feature. 
Figure~\ref{fig:low-calrity-heatmaps} presents the heatmap for feature 475, whose description emphasizes the occurrence of the token ``self.'' However, the feature does not activate on all instances of ``self'' (highlighted in yellow), indicating that a crucial aspect of the concept is missing or remains unclear. Additionally, the feature activates on other tokens, further suggesting that the description is incomplete.
This is reflected in the evaluation metrics: Clarity of 0.005 indicates that the concept is not expressed precisely enough for the evaluating LLM to generate synthetic data that reliably activates the feature. A Responsiveness score of 0.93 suggests that the feature does activate on natural data aligned with the concept, while a Purity score of 0.81 reveals that although the feature is primarily associated with the described concept, it also responds to other inputs.
Similar issues arise in features 653 and 821, where the underlying concepts appear highly specific -- activating on a particular token within a specific context. However, their descriptions are overly broad, making it difficult to generate synthetic data that reliably triggers the feature. 

This suggests that despite greater monosemanticity, interpreting SAE features remains challenging due to the difficulty in generating precise descriptions. In contrast to the Clarity metric, Responsiveness and Purity are on average higher for SAEs, as these metrics are less sensitive to imprecise descriptions and still align with the underlying concept. The higher Purity in SAEs aligns with our expectations of their greater monosemanticity.

\begin{table}[t]
\scriptsize
\centering
\begin{tabular}{p{1.8cm}p{0.9cm}p{0.9cm}p{0.9cm}p{0.9cm}} 
SAE Size        &   Clarity     & Respon-siveness & Purity         & Faithfulness     \\ \hline \hline 
MaxAct* 16K     & \textbf{0.57} &   \textbf{0.78} &  \textbf{0.69} &  \textbf{0.17}   \\
MaxAct* 65K     &    0.46       &   0.71          &  0.66          &  0.15            \\ \hline
Neuronpedia 16K & \textbf{0.43} &   \textbf{0.67} &  \textbf{0.60} &  \textbf{0.17}   \\
Neuronpedia 65K &    0.29       &   0.64          &  0.56          &  0.13            \\\hline 
\end{tabular}
\caption{Comparison results for SAEs of different sizes, see metrics distributions on Figure~\ref{fig:sae-sizes}. }
\label{tab:saes-sizes-comparison}
\end{table}
\paragraph{Interpreting larger SAEs is more difficult} 
We investigate whether SAEs with a higher number of features inherently exhibit a better alignment with the feature descriptions. To quantify this, we compare feature descriptions from \texttt{Gemma Scope 16K}  and   \texttt{Gemma Scope 65K}. We compare it based on Neuronpedia feature descriptions, as well as the ones obtained in this work via MaxAct*.
Consistent with our previous finding, our results indicate that increasing the number of SAE features does not inherently improve the alignment of features with their descriptions, as shown in Table~\ref{tab:saes-sizes-comparison}. We hypothesize that this stems from a finer-grained decomposition of concepts, making it more challenging for the explainer LLM to capture and articulate the precise concept. 

\begin{table}[t]
\scriptsize
\centering
\begin{tabular}{p{1.2cm}p{1cm}p{1cm}p{1cm}p{1cm}} 
Layer &   Clarity     & Respon-siveness & Purity        & Faithfulness \\ \hline \hline 
\noalign{\vskip .5mm}  
3              &    0.60       &   0.71          &  0.55         &  0.009   \\
12             &    0.44       &   0.54          &  0.43         &  0.016   \\
20             &    \textbf{0.67}       &   \textbf{0.74}          &  0.61         &  0.011   \\
25             &    0.59       &   0.65          &  0.54         &  \textbf{0.025}  \\ \hline 
\end{tabular}
\caption{Evaluation results for different layers in \texttt{Gemma-2}, see metrics distributions on Figure~\ref{fig:gemma2-layers}. }
\label{tab:gemma2-layers-comparison}
\end{table}

\paragraph{Interpretability varies across layers}
Table~\ref{tab:gemma2-layers-comparison} presents an evaluation of feature descriptions from different layers of \texttt{Gemma-2-2b}. 
Our analysis identifies layer 12 as the most challenging to interpret. A manual inspection of 50 randomly sampled features confirms these results: features in layer 12 exhibit high polysemanticity. 
The highest scores are observed in layer 20, with the exception of the Faithfulness metric. However, this may be due to the fine-tuning of the MaxAct* approach on this layer, which introduces a bias that specifically affects Faithfulness.
The highest Faithfulness score is observed in layer 25, while the lowest is found in layer 3.

\subsection{Evaluating Autointerpretability: Prompts, Examples, and Model Size}

Despite the growing number of methods proposed in automated interpretability research, there has been surprisingly little comprehensive evaluation of different approaches. In this section, we present a series of experiments that assess different components of feature generation pipelines and demonstrate how \ours\ can help in fine-tuning interpretability pipelines. 

\paragraph{Prompting with numerical- or delimiter-based input}
Prompt construction can significantly influence the quality of the generated descriptions. We investigate two primary approaches: passing (word, activation) pairs and using \{\{delimiters\}\} to highlight the most activated tokens (see Appendix~\ref{appendix-prompts} for more details). 
Our experiments indicate that the numerical input performs slightly better than the delimiter-based prompt, which contradicts previous research \cite{choi2024automatic}.

\begin{table}[t]
\scriptsize
\centering
\begin{tabular}{p{0.7cm}p{0.8cm}p{0.8cm}p{0.8cm}p{0.8cm}p{1.2cm}} 
Model                     & Input type          & Clarity          & Respon-siveness   & Purity        & Faithfulness \\ \hline \hline 
\multirow{2}{1cm}{Gemma-2} &  delimiters         &    \textbf{0.67}     &   0.74      &  0.61     & 0.01 \\ 
                          &  numeric            &    \textbf{0.67}     &   \textbf{0.76}      &  \textbf{0.64}     & 0.01  \\ \hline
\multirow{2}{1cm}{Gemma Scope}&  delimiters     &    0.57     &   0.78      &  0.69     & 0.16 \\  
                          &  numeric            &    \textbf{0.59}     &   \textbf{0.80}      &  \textbf{0.72}     & \textbf{0.17} \\ \hline 
\end{tabular}
\caption{Comparison results of activations input types via delimiters vs numerical input, see metrics distributions on Figure~\ref{fig:prompts_analysis} (a). }
\label{tab:numberical-vs-delimeters}
\end{table}

\paragraph{Few-shot prompting improves description quality}
Next we test how many examples should be passed to the Explainer Model in the prompt. We compare 0-shot (without examples), 1-shot, 2-shot, 5-shot, 10-shot and 20-shot prompts on \texttt{Gemma Scope}. In these variations, we use the delimiter-based prompts. The results, provided in Table~\ref{tab:number-of-shots-in-prompt}, demonstrate, that a larger number of examples brings steady improvement in Clarity, Responsiveness and Purity. Faithfulness shows no clear trend.  

\begin{table}[t]
\scriptsize
\centering
\begin{tabular}{p{1.2cm}p{1cm}p{1cm}p{1cm}p{1cm}} 
Number of shots &   Clarity     & Respon-siveness & Purity        & Faithfulness \\ \hline \hline 
0-shot             &    0.53       &   0.76          &  0.70         &  0.17   \\
1-shot             &    0.55       &   0.76          &  0.68         &  \textbf{0.19}  \\ 
2-shot             &    0.57       &   0.78          &  0.69         &  0.17  \\ 
5-shot             &    0.60       & 0.79            & 0.72           &  0.16  \\
10-shot             &   0.60       & 0.79            & 0.72           &  0.16  \\
20-shot             & \textbf{0.61} & \textbf{0.81}   & \textbf{0.73} &  0.16  \\ \hline 
\end{tabular}
\caption{Comparison results for different number of shots provided to the Explainer Model with the prompt based on Gemma Scope, see metrics distributions on Figure~\ref{fig:prompts_analysis} (c). }
\label{tab:number-of-shots-in-prompt}
\end{table}

\paragraph{Providing more samples increases evaluation scores}
We test 5, 15 and 50 samples, using the same delimiter-based prompts (see Table~\ref{tab:comparison-samples-number}). The results indicate, that increasing the number of samples improves description quality, though the gains are not substantial for any of the tested number of samples.

\begin{table}[t]
\scriptsize
\centering
\begin{tabular}{p{0.7cm}p{0.8cm}p{0.8cm}p{0.8cm}p{0.8cm}p{1.2cm}}  
Model & \# samples & Clarity      & Respon-siveness & Purity   & Faithfulness \\ \hline \hline 
\multirow{3}{1cm}{Gemma2} & 5      &    0.63 &  0.72   & 0.59   &  \textbf{0.01}    \\
                          & 15     &    0.67 &  0.74   & 0.61   &  \textbf{0.01}  \\ 
                          & 50     &    \textbf{0.69} &  \textbf{0.77}   & \textbf{0.62}   &  \textbf{0.01} \\  \hline 
\multirow{3}{1cm}{Gemma Scope} & 5 & 0.56          & 0.77          & 0.67          & 0.17 \\
                          & 15     & 0.57          & 0.78          & 0.69          & \textbf{0.18} \\ 
                          & 50     & \textbf{0.60} & \textbf{0.79} & \textbf{0.71} & 0.17 \\  \hline 
\end{tabular}
\caption{Comparison results for different number of activating samples provided to the Explainer Model, see metrics distributions on Figure~\ref{fig:prompts_analysis} (b). }
\label{tab:comparison-samples-number}
\end{table}

\paragraph{Better models produce better feature descriptions} 
Similarly to the experiment, comparing different Evaluation Models presented in Table~\ref{tab:evaluation-models-comparison}, we compare different Explainer Models and demonstrate the results in Table~\ref{tab:comparison-explainer-models}. 
\texttt{GPT-4o} achieves the highest scores, with \texttt{Llama-3.3-70B-Instruct} (AWQ 4-bit quantized) and \texttt{GPT-4o mini} as close alternatives. The smaller models struggle with assigning reasonable feature descriptions, in particular \texttt{Llama-3.2-1B}, which frequently fails to maintain a consistent response structure (see Appendix~\ref{appendix-extended-results}). 

\begin{table}[t]
\scriptsize
\centering
\begin{tabular}{p{1.6cm}p{1cm}p{1cm}p{1cm}p{1cm}} 
Model                 &   Clarity     & Respon-siveness & Purity        & Faithfulness \\ \hline \hline 
Llama-3.2-1B  &    0.39       &   0.56          &  0.35         &  0.10  \\  
Llama-3.2-3B  &    0.51       &   0.73          &  0.61         &  0.15  \\ 
Llama-3.1-8B  &    0.50       &   0.75          &  0.66         &  0.17   \\ 
Llama-3.3-70B &    0.54       &   0.78          &  0.70         &  \textbf{0.19}  \\ 
GPT-4o mini           &    0.58       &   0.78          &  0.70         &  0.17  \\ 
GPT-4o                &    \textbf{0.61}       &   \textbf{0.80}          &  \textbf{0.73 }        &  0.17  \\ \hline

\end{tabular}
\caption{Explainer models comparison, see metrics distributions on Figure~\ref{fig:explainer-models}. }
\label{tab:comparison-explainer-models}
\end{table}

\paragraph{Baselines fail in predictable ways}
To assess the effectiveness of MaxAct* approach, we compare it against established baseline methods, including the Neuronpedia feature descriptions, a TF-IDF \cite{ramos2003using, salton1988term} based description approach, and an unembedding-based method \cite{bloom2024understandingfeatureslogitlens} (see Appendix~\ref{appendix-autointerpretability-pipeline} for methodological details). The results are presented in Table~\ref{tab:baselines-gemmascope}.
MaxAct* consistently outperforms baselines in Clarity, Responsiveness, and Purity. Notably, the unembedding method achieves the highest Faithfulness score, a result that aligns with our expectations and related work \cite{gurarieh2025enhancingautomatedinterpretabilityoutputcentric}. Since this method explicitly considers the output that a given feature promotes, it naturally excels at capturing causal influence of the feature. However, this focus on output consistency often comes at the expense of Clarity, Responsiveness, and Purity, as raw unembedding-based descriptions do not incorporate any information about what activates the feature.
These findings again highlight the necessity of a holistic evaluation framework, as different methods optimize for different aspects of interpretability.

\begin{table}[t]
\scriptsize
\centering
\begin{tabular}{p{1.4cm}p{1cm}p{1cm}p{1cm}p{1cm}} 
Approach &   Clarity     & Respon-siveness & Purity        & Faithfulness \\ \hline \hline 
Neuronpedia  &    0.43       &   0.67          &  0.60         &  0.21   \\%
TF-IDF       &    0.42       &   0.72          &  0.53         &  0.21   \\%
Unembedding  &    0.38       &   0.65          &  0.61         &  \textbf{0.29}   \\%
MaxAct* &    \textbf{0.57}       &   \textbf{0.78 }         &  \textbf{0.69 }        &  0.21  \\ \hline 
\end{tabular}
\caption{Comparison of the quality of our feature descriptions to baselines, see metrics distributions on Figure~\ref{fig:baselines}. }
\label{tab:baselines-gemmascope}
\end{table}

%% file: sections/5_Conclusion.tex
\section{Conclusion}

In this work, we presented \ours, a new automated evaluation framework designed to rigorously evaluate the alignment between features and their open-vocabulary feature descriptions. By combining four complementary metrics \textit{Clarity}, \textit{Responsiveness}, \textit{Purity}, and \textit{Faithfulness}, our approach gives a comprehensive assessment of how a feature reacts to instances of the described concept, an evaluation of the description itself as well as the feature's causal role in the model’s outputs. Through extensive experiments across different feature types, layers, and description generation mechanisms, we demonstrated that methods relying on a single metric (e.g., simulation-based approaches) often give incomplete or misleading feature descriptions. Our framework can be used to highlight both the strengths and weaknesses of existing methods, while it also helps in debugging and improving these methods. We highlighted multiple results for improving the quality of feature explanations, such as using larger, more capable LLMs for the explainer and including more examples in the prompt. We hope that the open-source implementation of \ours\ will drive further research in automated interpretability and help make language models more transparent and safe to use.

%% file: sections/Limitations.tex
\section*{Limitations}
Despite presenting a comprehensive and robust evaluation framework, our work has certain limitations that we want to highlight here:
One key limitation when using LLMs is the inherent biases that can influence both rating and synthetic data generation, thus affecting our evaluation metrics. When rating concepts, an LLM's inherent bias may systematically lead to lower ratings for certain cultural contexts, domains, or languages outside its primary training distribution. Similarly, during synthetic data generation, LLMs may produce less diverse or representative data for underrepresented domains and low-resource languages. For example, an LLM might recognize a feature encoding a concept in English as directly representing that concept, whereas the same feature in another language might be classified with the additional specification of the language. This discrepancy could also lead to unintended biases when steering models based on these interpretations. 

Similar issues may arise from biases present in the datasets used in the evaluation procedure. While we used The Pile dataset due to its extensive nature, underrepresented concepts may still lead to less stable purity and responsiveness measures. Although FADE can help flag such cases and enable us to achieve fuller coverage by identifying the aforementioned training dataset issues, ultimately a more balanced corpus would be beneficial. 

Another limitation is related to the steering behavior in the Faithfulness pipeline. Our current implementation does not explicitly verify whether the generated sequences under modification remain grammatically correct and semantically meaningful. However, this can be mitigated in the future by integrating KL-divergence to control the magnification factor, used for steering the features \cite{paulo2024automaticallyinterpretingmillionsfeatures, gurarieh2025enhancingautomatedinterpretabilityoutputcentric}. 
Finally, our Faithfulness measure is not well-suited for capturing the behavior of inhibitory neurons. While a neuron may causally suppress the presence of a concept in the model’s output, our metric, by design, is limited in detecting decreases in the activation of sparse concepts. Moreover, current automated interpretability methods, which our evaluation framework is intended to support, typically do not focus on identifying inhibitory neurons. However, if an inhibitory feature is redefined in affirmative terms (e.g., instead of "the feature suppresses English," phrased as "the feature promotes non-English languages"), our Faithfulness metric can handle such cases appropriately. Nonetheless, a more comprehensive investigation into inhibitory neurons, both in terms of automated interpretability and evaluation, remains necessary.

%% file: sections/Acknowledgements.tex
\section*{Acknowledgements}
We sincerely thank Melina Zeeb for her valuable assistance in creating the graphics and logo for {\ours}. This work was supported by the Federal Ministry of Education and Research (BMBF) as grant BIFOLD (01IS18025A, 01IS180371I); the European Union’s Horizon Europe research and innovation programme (EU Horizon Europe) as grants [ACHILLES (101189689), TEMA (101093003)]; and the German Research Foundation (DFG) as research unit DeSBi [KI-FOR 5363] (459422098).

%% file: sections/Appendix.tex
\label{sec:appendix}
\section{Extended related work}
\label{appendix-related-work}

A common approach to automatic interpretability includes selecting data samples that strongly activate a given neuron and using these samples, along with their activations as input to a larger LLM, that serves as an Explainer Model and generates feature descriptions \cite{bills2023language, bricken2023monosemanticity, choi2024automatic, paulo2024automaticallyinterpretingmillionsfeatures, rajamanoharan2024jumpingaheadimprovingreconstruction}. Previous research has investigated various factors influencing this method, including prompt engineering, the number of data samples used, and the size of the Explainer Model (see Appendix~\ref{appendix-related-work}).

Building on this approach, \cite{choi2024automatic} advanced the method by fine-tuning \texttt{Llama-3.1-8B-Instruct} on the most accurate feature descriptions, as determined by their  simulated-activation metric. This fine-tuning aimed to improve the performance and accuracy of description generation, ultimately outperforming \texttt{GPT-4o} mini.

An output-centric approach was introduced by
\cite{gurarieh2025enhancingautomatedinterpretabilityoutputcentric} in an  attempt to address another key challenge -- the feature descriptions generated via the data samples that activate the feature the most, often fail to reflect its influence on the model’s output. The study demonstrates that a combined approach, integrating both activation-based and output-based data, results in more accurate feature descriptions and improves performance in causality evaluations.

Several studies perform their experiments exclusively on SAEs \cite{goodfire, paulo2024automaticallyinterpretingmillionsfeatures, rajamanoharan2024jumpingaheadimprovingreconstruction,templeton2024scaling}, while others focus on MLP neurons \cite{bills2023language, choi2024automatic}. Although \cite{templeton2024scaling} compares the interpretability of SAEs to that of neurons and concludes that features in SAEs are significantly more interpretable, these findings heavily rely on qualitative analyses. 

Previous research consistently shows that open-source models are effective for generating explanations, with advanced models producing better descriptions. For instance, \citet{bills2023language} report that GPT-4 achieves the highest scores, whereas Claude 3.5 Sonnet performs best for \citet{paulo2024automaticallyinterpretingmillionsfeatures}. The number of data samples used to generate feature descriptions also vary across studies. \citet{bills2023language} use the top five most activating samples, while \citet{choi2024automatic} select 10-20 of the most activating samples to generate multiple descriptions for evaluation. In contrast, \citet{paulo2024automaticallyinterpretingmillionsfeatures} use 40 samples and suggest that randomly sampling from a broader set of activating leads to descriptions that cover a more diverse set of activating examples, whereas using only top activating examples often yields more concise descriptions that fail to capture the entire description.

The datasets used in these studies also differ. \citet{paulo2024automaticallyinterpretingmillionsfeatures} utilize the RedPajama 10M \cite{together2023redpajama} dataset, \citet{choi2024automatic} use the full LMSYSChat1M \cite{zheng2023lmsys} and 10B token subset of FineWeb \cite{penedo2024the}, \citet{bills2023language} -- WebText\cite{radford2019language} and the data used to train GPT-2 \cite{radford2019language}. Additionally, different delimiter conventions are observed: \citet{choi2024automatic} use {} delimiters, \citet{paulo2024automaticallyinterpretingmillionsfeatures} use << >>, and \citet{bills2023language} use numerical markers.

\section{Data Preprocessing}
\label{appendix-dataset}
For our work, we used an uncopyrighted version of the Pile dataset, with all copyrighted content removed, available on Hugging Face \cite{gao2020pile} (\url{https://huggingface.co/datasets/monology/pile-uncopyrighted}). This version contains over 345.7 GB of training data from various sources. From this dataset, we extracted approximately 6 GB for labelling while preserving the relative proportions of the original data sources. The extracted portion from the training partition was used to collect the most activated samples. For evaluations, we utilized the test partition from the same dataset, applying identical preprocessing steps as those used for the training data.
    
\begin{table}[h!]
\scriptsize
\centering 
\setlength{\tabcolsep}{6pt}  
\begin{tabular}{@{}lrr@{}}  
Component & Size (GB) & Proportion (\%) \\  \hline \hline
Pile-CC             & 1.93  & 32.17 \\
PubMed Central     & 1.16  & 19.38 \\
ArXiv             & 0.70  & 11.67 \\
FreeLaw          & 0.55  &  9.14 \\
PubMed Abstracts    & 0.32  &  5.39 \\
USPTO Backgrounds & 0.31  &  5.23 \\
Github         & 0.27  &  4.54 \\
Gutenberg (PG-19)  & 0.23  &  3.85 \\
Wikipedia (en)      & 0.15  &  2.57 \\
DM Mathematics      & 0.11  &  1.87 \\
HackerNews          & 0.07  &  1.17 \\
Ubuntu IRC          & 0.06  &  0.99 \\
EuroParl            & 0.06  &  0.95 \\
PhilPapers          & 0.03  &  0.58 \\
NIH ExPorter        & 0.03  &  0.50 \\
\hline
\textbf{Total}             & \textbf{5.99} & \textbf{100.00} \\ \hline
\end{tabular}
\caption{Extracted dataset and proportion of sub components}
\label{tab:dataset}
\end{table}

Our preprocessing involved several steps to ensure a balanced and informative dataset. First, we used the NLTK \cite{bird2009natural} sentence tokenizer to split large text chunks into individual sentences. We then filtered out sentences in the bottom and top fifth percentiles based on length, as these were typically out-of-distribution cases consisting of single words, characters, or a few outliers. This step helped achieve a more balanced distribution. Additionally, we removed sentences containing only numbers or special characters with no meaningful content. Finally, duplicate sentences were deleted.

\begin{table}[h]
\scriptsize
\centering
\setlength{\tabcolsep}{6pt}  
\begin{tabular}{@{}lrr@{}}  
\textbf{} & Labeling Dataset & Evaluation Dataset \\
\hline \hline
Number of sentences & 88{,}689{,}425 & 5{,}443{,}427 \\
Number of tokens & 2{,}284{,}636{,}243 & 137{,}600{,}815 \\
Number of unique tokens & 21{,}707{,}092 & 2{,}336{,}552 \\
\hline
\end{tabular}
\caption{Dataset Statistics}
\label{tab:dataset-stats}
\end{table}

\section{Automated Interpretability Pipeline}

\subsection{Implementation of Automated Interpretability}
\label{appendix-autointerpretability-pipeline}

\begin{table*}[t]
\scriptsize
\centering
\begin{tabular}{p{2.2cm}p{1.2cm}p{4.6cm}p{6cm}} 
Model                 &   Layers       &       HuggingFace             & Descriptions   \\ \hline \hline 
Gemma-2-2b            & 3, 12, 20, 25 &   google/gemma-2-2b  &  --     \\
Gemma Scope 16K       &    20         & google/gemma-scope-2b-pt-res/tree/main/ layer\_20/width\_16k/average\_l0\_71  & neuronpedia.org/gemma-2-2b/20-gemmascope-res-16k    \\
Gemma Scope 65K       &    20         & google/gemma-scope-2b-pt-res/tree/main/ layer\_20/width\_65k/average\_l0\_114  & neuronpedia.org/gemma-2-2b/20-gemmascope-res-65k    \\
Llama3.1-8B-Instruct  & 19   & meta-llama/Llama-3.1-8B-Instruct &  github.com/TransluceAI/observatory.git \\ \hline 
\end{tabular}
\caption{Sources for models, SAEs and feature descriptions, used in this work. }
\label{tab:models-sources}
\end{table*}

Our experiments are based on models and feature descriptions presented in Table~\ref{tab:models-sources}. We generate our feature descriptions as follows. 
In the implementation of the autointerpretability pipeline, we are closely following \cite{bills2023language, paulo2024automaticallyinterpretingmillionsfeatures, rajamanoharan2024jumpingaheadimprovingreconstruction} and others: we pass the natural dataset, used for generating descriptions (see Appendix~\ref{appendix-dataset}) through the model, and via forward hooks we access the activations of each feature and each layer, which can be done in parallel. For \texttt{Gemma Scope} SAEs we implement a wrapper class that is built into the model as named module, and can be easily extended to other SAEs.

After passing the whole dataset through the model, we take top $10^3$ data samples for each feature, based on the maximum activation of tokens. We only consider a single maximum activating token of a data sample. Later, we uniformly subsample the necessary number of data samples, i.e. 5, 15, ..., 50, that are further passed to an explainer LLM. We do that to avoid outliers -- a small part of the data which activates the feature the most, but does not represent its overall behavior well. The considered top $10^3$ most activating samples represent top 0.001\% of the dataset. More complex sampling strategies can yield better performance, as described in Appendix~\ref{appendix-related-work}, but their implementation and evaluation is left out for future work. 

Experiments are performed on the following models: \texttt{Gemma-2-2b} layer 20 \cite{gemmateam2024gemma2improvingopen}, \texttt{Llama-3.1-8B-Instruct} layer 19 \cite{grattafiori2024llama3herdmodels}, and SAEs \texttt{Gemma Scope} 16K and 65K layer 20 \cite{lieberum-etal-2024-gemma}. 
In addition, we generate descriptions using baseline methods, namely TF-IDF and unembedding matrix projection. \textit{Term Frequency-Inverse Document Frequency (TF-IDF)} is a widely used technique in NLP for measuring the importance of a word in a document relative to a corpus. It balances word frequency with how uniquely the word appears across documents, assigning higher scores to informative words while down-weighting common ones. We generate these values using 15 maximally activating samples. On the other hand, the \textit{unembedding matrix} (\( W_U \)) \cite{bloom2024understandingfeatureslogitlens} in transformer models maps the residual stream activations to vocabulary logits, determining word probabilities in the output. By analyzing projections onto this unembedding matrix, we gain insight into how learned features influence token predictions. 
To generate SAE unembedding descriptions, we generate a logit weight distribution across the vocabulary, and then use the top ten words with the highest probabilities as feature descriptions.
\[
\text{Logit weight distribution} = W_U*W_{dec}[feature]
\]
where \( W_U \) is the unembedding matrix of a transformer model and \( W_{dec}\) are the decoder weights of sparse auto encoders.

This reveals which words are most associated with a given feature, enabling interpretability of sparse autoencoder (SAE) features, as well as MLP neurons. Both these methods are cheap baselines to compare with descriptions generated via different autointerpretability approaches.

\subsection{Prompt Engineering}
\label{appendix-prompts}
Our feature description pipeline consists of several key components, such as the Subject Model, for which the descriptions are generated, the Explainer Model, which is a larger language model, used to generate descriptions, and the System Prompt, which provides task-specific instructions to the Explainer Model, detailing what to focus on in the provided samples and how to format the output. We append 5, 10, or 50 sentences along with a \texttt{user\_message\_ending} at the end, which helps reinforce the expected output structure. Before normalizing activations, we first average activations for tokens belonging to the same word. These values are then normalized between 0 and 10. For delimiters, we use single curly brackets if the activation intensity is below 4, and double curly brackets otherwise. For numerical input we provide the Explainer Model with a dictionary of the most activated tokens after each sample. 
\begin{table}
    \scriptsize
    \centering
    \begin{tabular}{lcc}
        & Avg \# of tokens & Cost/$10^3$ feature (\$)\\
        \hline
        \hline
        \noalign{\vskip .5mm}  
 0-shot &  1,423 & 2.13\\
 1-shot &  1,498 & 2.25\\
        2-shot  & 1,564  & 2.35\\
        10-shot & 2,395  & 3.95\\
        20-shot & 3,393  & 5.09\\
        \hline
    \end{tabular}
    \caption{Token usage and cost comparison for different shot settings.}
    \label{tab:token-cost-comparison}
\end{table}

\begin{table}
    \scriptsize
    \centering
    \begin{tabular}{cc}
        \# of sentences & Avg \# of tokens \\
        \hline
        \hline
        \noalign{\vskip .5mm}  
        5  & 333  \\
        15 & 964  \\
        50 & 2,507 \\
        \hline
    \end{tabular}
    \caption{Average number of tokens as the number of sentences increases. These values are based on tokenizer used for \texttt{Gemma-Scope-16k} and are not the number of tokens per request generated by OpenAI.}
    \label{tab:avgtokens_per_numberofsamples}
\end{table}
\paragraph{Main prompt:}
The main prompt we use to generate descriptions is given below. 

\begin{lstlisting}

You are a meticulous AI researcher conducting an important investigation into sparse autoencoders of a language model that activates in response to specific tokens within text excerpts. Your overall task is to identify and describe the common features of highlighted tokens, focusing exclusively on the tokens that activate and ignoring broader sentence context unless absolutely necessary.

You will receive a list of sentences in which specific tokens activate the neuron. Tokens causing activation will appear between delimiters like {{ }}. The activation values range from 0-10:

- If a token activates with an intensity of <4, it will be delimited like {{this}}.
- If a token activates with an intensity of >4, it will be delimited like {{{{this}}}}.

Guidelines:
1. Focus on the activated tokens: The description must primarily relate to the highlighted tokens, not the entire sentence.
2. Look for patterns in the tokens: If a specific token or a group of similar tokens repeatedly activates, center your analysis on them.
3. Sparse Autoencoder Dependency: The activations depend only on the words preceding the highlighted token. Descriptions should avoid relying on words that come after the activated token.
4. No coherent presence of concept: Return 'NO CONCEPT FOUND' if there is no coherent theme in the sentences provided, do not force a concept and stay grounded.
Output Format:
Concept: [Focus on the common concept tied to the highlighted tokens, described in a concise phrase.]

### Example:
Input example 1:
Sentence 1: The {{United States}} will not allow {{threats}} against its people.  
Sentence 2: The {{U.S.}}  emphasizes deterrence in foreign policy.

Example Output:  
Concept: U.S. deterrence policies and moral stance on global threats.

Input example 2:
Sentence 1:  Ash/Brock [Bouldershipping]\nI forget about this one {{all}} for one favours.
Sentence 2: I see this attitude brewing {{all}} at once.
Sentence 3: I get emails from people {{all}} over the world.

Example Output:  
Concept: Presence of the word 'all'.

user_message_ending: >
Analyze all these sentences as ONE corpus and provide your description in the following format:

Concept: Your devised concept and it's description in a concise manner, and very few words.

\end{lstlisting}

\paragraph{Prompt variation 1:} Main prompt + dictionary of most activated tokens with their normalized activation values. 
\begin{lstlisting}

You are a meticulous AI researcher [...]
 
You analyze a list of most-activated sentences and a dictionary of relevant tokens, each with assigned activation values, to identify key themes and concepts. Tokens causing activations will be provided in the dictionary after each sentence. The activation values range from 0-10.
          
Guidelines:[..]
          
Output Format:
Concept: [Focus on the common concept tied to the highlighted tokens, described in a concise phrase.]
          
EXAMPLE 1
Sentence 1: "The choir's harmonies resounded throughout the church as the congregation stood in awe."
Most relevant tokens: {{"harmonies": 9, "resounded": 8, "church": 6, "congregation": 4}}[..]

\end{lstlisting}
\textbf{Prompt variation 2:} Main prompt + zero shots. \newline
\textbf{Prompt variation 3:} Main prompt + one shot.  \newline
\textbf{Prompt variation 4:} Main prompt + five shots.  \newline

\subsection{Computational Costs}
\label{appendix-autointerpretability-costs}

Obtaining maximum activating samples for features is performed locally on a cluster with 8 NVIDIA A100 GPUs with VRAM 40Gb. The process can be easily parallelized, but for cost calculation we are using the total consumed time. Such GPUs can be rented starting with \$0.67/hr, however, an average price on the market exceeds this value. For simplicity of calculations, we take a price of \$1/hr. For convenience, we consider a price per $10^3$ features, the results are presented in Table~\ref{tab:maxact-costs}. 

\begin{table}[t]
    \scriptsize
    \centering
    \begin{tabular}{p{1.9cm}p{1.2cm}p{0.85cm}p{2.15cm}}
                             & \# of features & Time (h) & Cost/$10^3$ features (\$) \\
        \hline
        \hline
        \noalign{\vskip .5mm}  
        Gemma-2-2b            & 239,616  & 185  & 0.77  \\
        Gemma Scope 16K       & 16,384  & 524  & 31.98  \\
        Gemma Scope 65K       & 65,536  & 622  & 9.49 \\
        Llama-3.1-8B          & 458,752  & 361  & 0.79  \\
        \hline
    \end{tabular}
    \caption{Cost comparison for discovering samples, that are maximally activating features. }
    \label{tab:maxact-costs}
\end{table}

Due to the differences in the implementation, calculating maximum activating samples is significantly cheaper for the complete models, since we do it during the same forward pass for the whole model, and the price is divided by the total number of features we consider. For SAEs, in the current implementation we only do it per one layer, which significantly reduces the number of considered features. In the future we are planning on optimizing this process in a way, that it is possible to calculate it simultaneously for multiple chosen SAEs. The size of SAEs also plays an important role: in terms of total costs it is more expensive, but if we consider the cost per $10^3$ features, the larger the SAE -- the cheaper it is. 

The total computational cost for this part of the experiments is estimated at 4,920 GPU hours. This includes all experiments conducted on the models referenced in this paper. While not all experiments are explicitly discussed, they contributed to the final results presented here.

\section{\ours \ Evaluation Framework}\label{appendix-evaluation}

\subsection{Implementation and Computational Efficiency}\label{appendix-evaluation-implementation}

To compute the Purity and Responsiveness, we sample evaluation sequences from the natural dataset based on the activation distribution of the considered feature. For each sequence, we calculate the maximum absolute activation value across all tokens. Using these values, we sample sequences by selecting a user-configurable percentage of those with the strongest activations and for the remainder, drawing an equal number of samples from each of the following percentile ranges: 
\([0\%, 50\%)\), 
\([50\%, 75\%)\), 
\([75\%, 95\%)\), 
and \([95\%, 100\%]\).

Computational efficiency is a key consideration in our design, as evaluating every neuron in an LLM can be prohibitively expensive. The cost of evaluations is dynamically adjustable based on several factors, including the number of samples generated and rated, the evaluating LLM used, and the natural dataset selection. 

Our method allows users to control the cost by setting the number of synthetic samples (denoted $n$) relative to the full size of the natural dataset ($N$). By pre-computing activations from the natural dataset in parallel, we effectively reduce the per-run complexity from $\mathcal{O}(N*M)$ for $M$ neurons to $\mathcal{O}(n*M)$. Given that $n$ is typically in the hundreds while $N$ is in the millions, this strategy yields significant efficiency gains.

Additionally, we only execute the computationally costly Faithfulness evaluation when both Clarity and Responsiveness exceed a user-configurable threshold. This conditional execution ensures that unnecessary computations are avoided for features that do not meet our interpretability criteria.

\subsection{Details on the Experiment Setup}\label{appendix-evaluation-experiment-setup}

In our experiments we send 15 requests to the evaluation LLM for generating synthetic samples. We remove duplicates and use these as concept-samples. We use the whole evaluation dataset as control samples. For rating we draw 500 samples from the natural dataset, where we take 50 from the top activated and 450 from the lower percentiles, according to the sampling strategy outlined above in Appendix~\ref{appendix-evaluation-implementation}. If we obtain fewer than 15 concept-samples in this first rating we again sample 500 new samples with the same sampling approach. We rate 15 samples at once, and if one of the calls fails, for example due to formatting errors of the evaluating LLM, we retry the failed samples once. 
For the Faithfulness experiments we use the modification factors $[-50, -10, -1, 0, 1, 10, 50]$. We draw 50 samples from the natural dataset and let the subject LLM continue them for 30 tokens. We then rate only these continuations for concept strength by the evaluating LLM, again retrying once, if the rating fails. We only execute the Faithfulness experiment, if both the Clarity and Responsiveness of a feature is larger or equal to 0.5. 

\paragraph{Licenses} \texttt{Gemma-2-2b} is released under a custom Gemma Terms of Use. \texttt{Gemma Scope} SAEs are released under Creative Commons Attribution 4.0 International. \texttt{Llama3.1-8B-Instruct} is released under a custom Llama 3.1 Community License. Transluce feature descriptions, Pile Uncopyrighted dataset and LangChain are released under MIT License. vLLM is released under  Apache 2.0 License.

\subsubsection{Prompts for the Evaluating LLM}\label{appendix-evaluation-experiment-setup-prompts}

\textbf{Generating Synthetic Data Prompt}

\begin{lstlisting}
You are tasked with building a database of sequences that best represent a specific concept. 
To create this, you will generate sequences that vary in style, tone, context, length, and structure, while maintaining a clear connection to the concept. 
The concept does not need to be explicitly stated in each sequence, but each should relate meaningfully to it. Be creative and explore different ways to express the concept.

Here are examples of how different concepts might be expressed:

Concept: "German language" - Sequences might include German phrases, or sentences.
Concept: "Start of a Java Function" - Sequences might include Java code snippets defining a function.
Concept: "Irony" - Sequences might include ironic statements or expressions.

Provide your sequences as strings in a Python List format.

Example: ["This is a first example sequence.", "Second example sequence but it is much longer also there are somy typos in it. wjo told you that I can type?"]

Output only the Python List object, without any additional comments, symbols, or extraneous content.
\end{lstlisting}

\textbf{Rating Natural Data Prompt}

\begin{lstlisting}
You are tasked with building a database of sequences that best represent a specific concept. 
To create this, you will review a dataset of varying sequences and rate each one according to how much the concept is expressed.

For each sequence, assign a rating based on this scale:

0: The concept is not expressed.
1: The concept is vaguely or partially expressed.
2: The concept is clearly and unambiguously present.

Use conservative ratings. If uncertain, choose a lower rating to avoid including irrelevant sequences in your database. 
If no sequence expresses the concept, rate all sequences as 0.

Each sequence is identified by a unique ID. Provide your ratings as a Python dictionary with sequence IDs as keys and their ratings as values.

Example Output: {{"14": 0, "15": 2, "20": 1, "27": 0}}

Output only the dictionary - no additional text, comments, or symbols."    
\end{lstlisting}

\subsubsection{Associated Cost}

The activation generation for the Subject Models for the results in section \ref{sec:Results} was run locally on a cluster with NVIDIA A100 GPUs with 40GB of VRAM. Similar to section \ref{appendix-autointerpretability-costs} we assume a price of \$ 1/hr. 
Since activations can be cached in parallel for the whole model, only a single pass over the evaluation dataset was needed per model. We estimate a needed time of 24 hours on one GPU for the activation generation, resulting in a cost of \$24 per model. 
During the evaluations, only the activations of synthetic samples and Faithfulness text generations need to be computed. Given our hyper-parameter setting described in D.2, we measure an average of 31 seconds GPU time per neuron. Of that, roughly 2 seconds are spent on the Clarity, Responsiveness and Purity metrics, while the much more expensive Faithfulness computation contributes the remaining 29 seconds. These 29 seconds arise because, with our present threshold of 0.5, the Faithfulness experiment is executed on 53\% of candidate labels; if it were executed every time it would add roughly 55 seconds. This leads us to an average cost of \$0.0086 per neuron, excluding the fixed activation caching cost.

The estimated cost for the evaluating LLM consists of the cost for the generation of synthetic samples as well as the cost for rating natural data. In the configuration used for the experiments, unless stated otherwise, we estimate the evaluating LLM creates on average about 2800 tokens per feature evaluation, which corresponds to a cost of about \$0.00168 per feature at an output token cost of \$0.6 per million tokens. For the rating part, the evaluating LLM receives about 23000 tokens as input, of which on average 17000 are from the Responsiveness and Purity experiments and 6000 are from the Faithfulness experiments, which corresponds to a total cost of \$0.00345 per feature at an input token cost of \$0.15 per million tokens. Corresponding results are presented in Table~\ref{tab:eval-subject-costs}. 

\begin{table}[t]
    \scriptsize
    \centering
    \begin{tabular}{p{1.9cm}p{.95cm}p{.85cm}p{.8cm}p{1.05cm}}
                             & Features evaluated & GPU Time (h) & Inference Cost (\$) & Cost per feature (\$) \\
        \hline
        \hline
        \noalign{\vskip .5mm}  
        Gemma-2-2b            & 9.000  & 24+77.5  & 46.17 & 0.0164 \\
        Gemma Scope 16K       & 25.000  & 24+215.3 & 128.25  & 0.0147  \\
        Gemma Scope 65K       & 2.000  & 24+17.2 & 10.26 & 0.0257 \\
        Llama-3.1-8B          & 2.000  & 24+17.2 & 10.26 & 0.0257 \\
        \hline
    \end{tabular}
    \caption{Computational cost estimates of the evaluation experiments.}
    \label{tab:eval-subject-costs}
\end{table}

The total cost in our experiments is primarily driven by the use of high values of hyperparameters intended to ensure high evaluation accuracy as outlined in section \ref{appendix-evaluation-experiment-setup}. Here especially the low threshold of executing the Faithfulness experiments is a big contributor to overall cost. Increasing this threshold will reduce the computational cost substantially. Additionally, the high number of samples that is being rated for the activation-based metrics can be reduced in order to significantly lower the amount of needed inference tokens for the evaluating LLM. Furthermore, as seen in Table \ref{tab:evaluation-models-comparison}, often the size of the evaluating LLM itself can be reduced with minimal accuracy losses, resulting in cheaper inference costs via e.g. self-hosted models or simply cheaper inference. For example if we had run the Faithfulness experiments only on the most promising 25\% percent of labels, corresponding to a Faithfulness threshold of approximately 0.9 given our evaluated labels, the average GPU compute cost per neuron, excluding the static computation of cached activations, would decrease from \$0.0086 to \$0.0044.

In our experiments we chose these high hyperparameter settings, as we wanted to prioritize robustness and replicability of our results. We believe however, that even with significantly lower configurations, especially regarding the Faithfulness-threshold and number of modification factors, valuable insights can be found at a considerably lower cost.

\subsection{Combining \ours\ metrics into a single score} The principal value of \ours\ lies in its multidimensional analysis of feature-to-description alignment. Reducing these dimensions to a single score may obscure important nuances that are crucial for developing and refining interpretability techniques. Separate metrics allow researchers to pinpoint specific weaknesses in their descriptions and address them directly. However, for applications that require simplified comparison or ranking of descriptions, appropriate combination methods can be selected based on the specific requirements of the use case.
We propose that various means (weighted, geometric, or harmonic) could serve the purpose of creating one holistic evaluation metric. A weighted combination of the four metrics would allow one to adjust importance based on specific use cases and applications. The geometric mean would be particularly suitable, as it accounts for all metrics while being naturally sensitive to underperforming dimensions. Alternatively, a harmonic mean could be employed for cases where balanced performance is even more critical, as it penalizes low scores more severely than the geometric mean.

\section{Extended Results}
\subsection{Quantitative Analysis}
\begin{figure*}[t]
    \centering
    \includegraphics[width=1\linewidth]{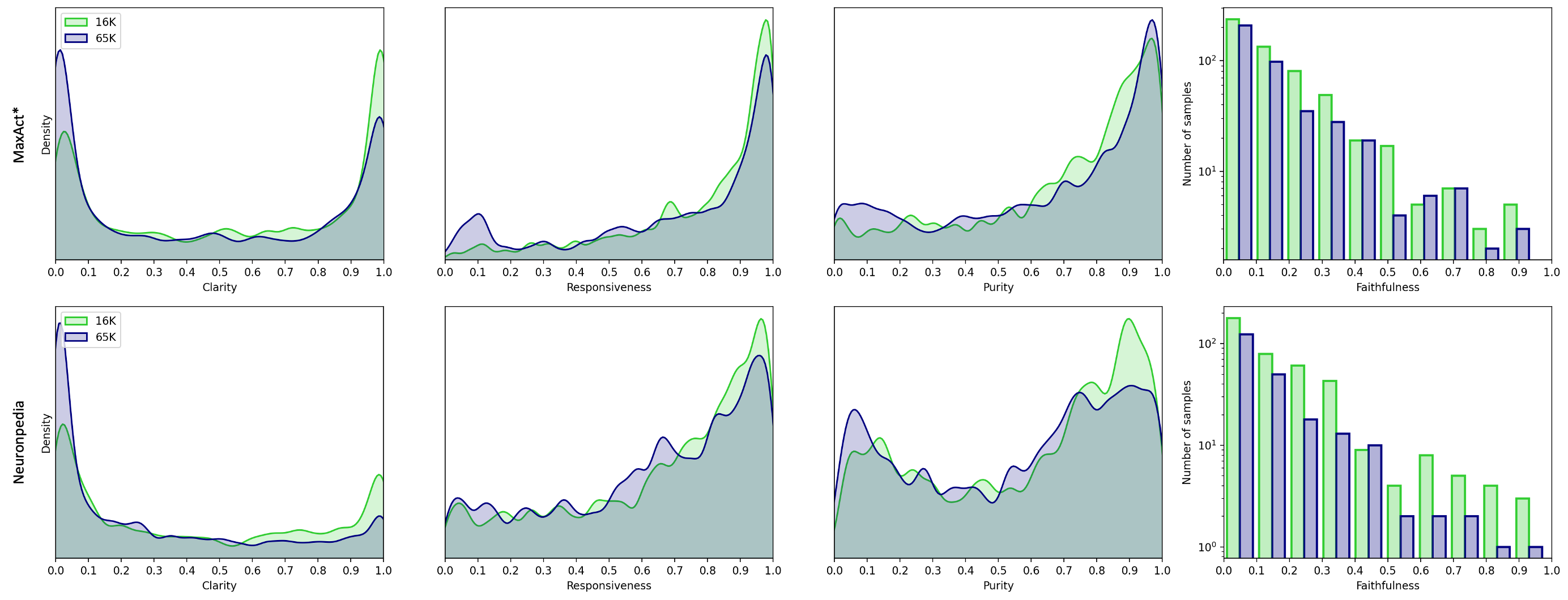}
    \caption{Feature descriptions fit for SAEs of different sizes.}
    \label{fig:sae-sizes}
\end{figure*}

\begin{figure*}[t]
    \centering
    \includegraphics[width=1\linewidth]{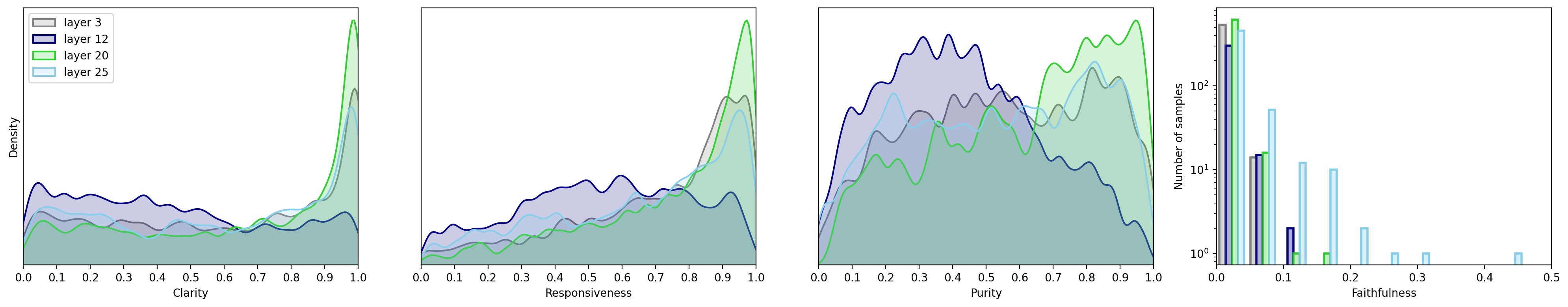}
    \caption{Feature descriptions fit for different layers of Gemma-2. }
    \label{fig:gemma2-layers}
\end{figure*}
\label{appendix-extended-results}

This section provides a more in-depth analysis of the experimental results presented in Section~\ref{sec:Results}.

\paragraph{SAEs concept narrowness results in lower interpretability} 
This might seem counterintuitive at first, but it is important to stress again that we are not evaluating the features by themselves, but instead the adequacy of the proposed feature description to these features. 
One potential explanation is that larger SAEs distribute concepts more finely across features. As a result, a slightly inaccurate description that might have still activated a feature in a smaller SAE may fail to activate the corresponding feature in a larger SAE, where concepts are encoded even more sparsely. The consistency of this result is demonstrated by using different feature descriptions -- MaxAct*, produced in this work, and the ones available on Neuronpedia. 

Interestingly, we observe a small left-skewed peak in the Responsiveness distribution for 65K SAEs labelled via MaxAct*, a pattern not seen in any other experiment (see Figure \ref{fig:sae-sizes}). A qualitative analysis suggests that this is primarily caused by features representing out-of-distribution concepts relative to the dataset used for evaluation, such as, e.g., ``new line'' feature, which due to the preprocessing steps is not present in the dataset used for descriptions generation (see Figure~\ref{fig:new-line-example}). Due to the lower quality of feature descriptions, the Purity distribution for Neuronpedia descriptions of \texttt{Gemma Scope 65K} exhibits a bimodal pattern. High-quality descriptions tend to have high Purity, reflecting the greater monosemanticity of the features. However, a substantial number of inaccurate descriptions lead to very low Purity values.

\begin{figure*}[t]
    \centering
    \includegraphics[width=1\linewidth]{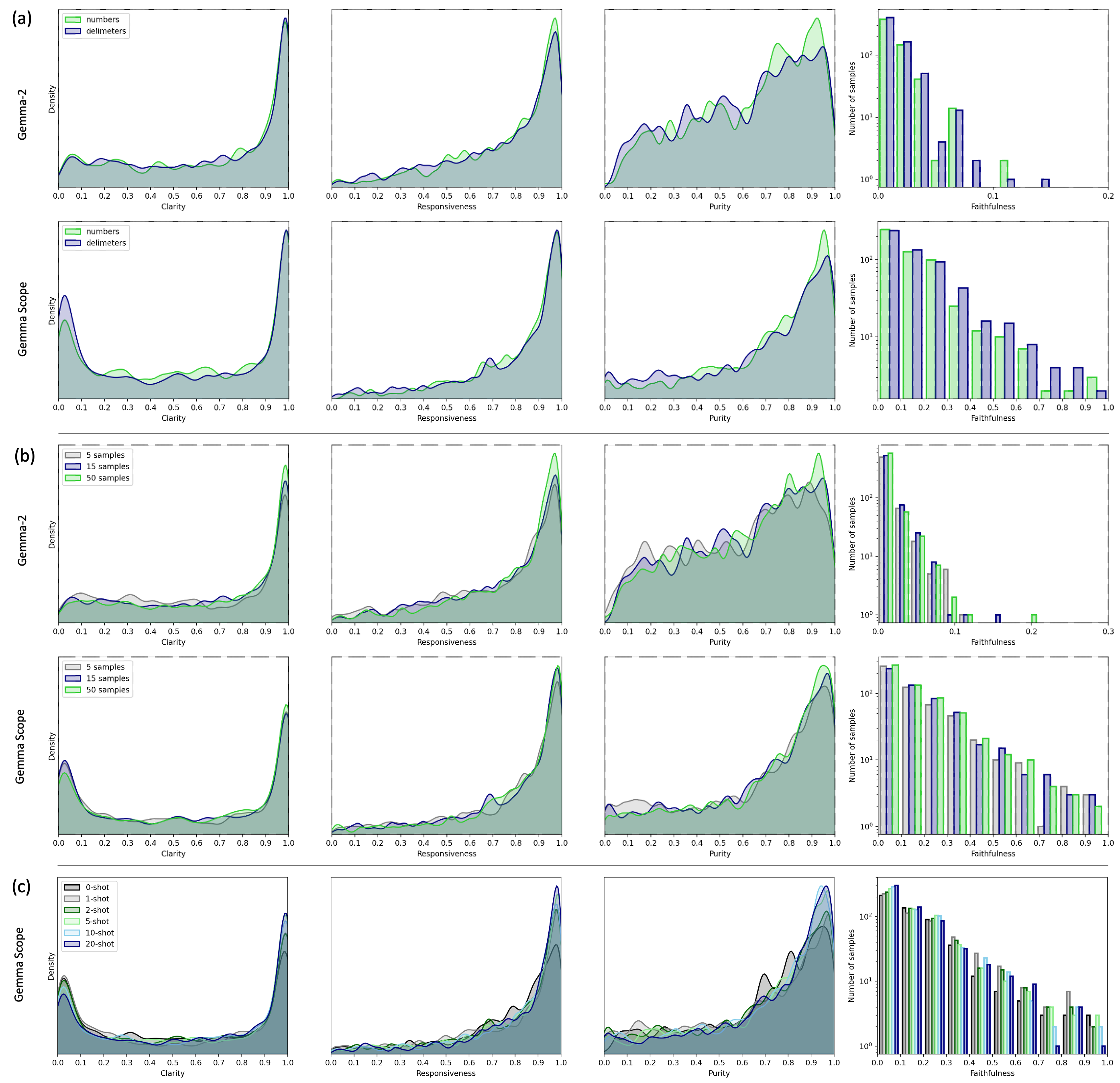}
    \caption{Analysis of the main prompt's components: (a) numeric input vs delimeter-based; (b) number of activating samples, provided to the explainer LLM; (c) number of shots used in the prompt. }
    \label{fig:prompts_analysis}
\end{figure*}

\paragraph{Some Gemma-2 layers look almost not interpretable} 
The feature distribution in layer 12 differs significantly from other layers, since it lacks the characteristic rightward elevation in Clarity, Responsiveness, and Purity scores, as presented on Figure~\ref{fig:gemma2-layers}. Manual analysis of the heatmaps support the theory, that layer 12 demonstrates a high level of polysemanticity. Interestingly, similar result were demonstrated on \texttt{Llama-3.1-8B-Instruct} by \cite{choi2024automatic}. Evaluating several complete models with \ours\ would provide more insights into the interpretability of different models components. 

\paragraph{Numeric input shows marginally better performance} As illustrated in Figure~\ref{fig:prompts_analysis} (a), the performance difference between numeric input-based prompts and delimiter-based highlighting is not statistically significant for both MLP neurons and SAEs, though the mean score is slightly higher for numeric input. A more comprehensive evaluation across multiple layers and a larger feature set is needed to determine the optimal approach. Nonetheless, these findings challenge prior assertions that highlighting the most activating tokens is superior due to LLMs' assumed difficulty in processing numerical inputs \cite{choi2024automatic}.

\paragraph{Increasing sample count improves performance} A clear trend emerges: providing more samples to the explainer LLM enhances its performance. However, this also increases the computational cost of feature generation due to the higher token count. While 15 samples yield strong results, 50 samples perform even better, as shown in Figure~\ref{fig:prompts_analysis} (b).

\begin{figure*}
    \centering
    \includegraphics[width=1\linewidth]{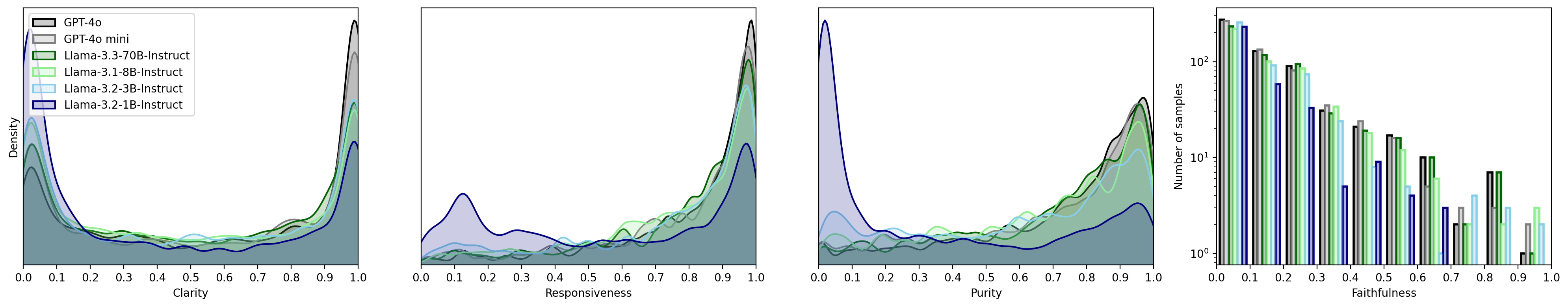}
    \caption{Performance of feature descriptions generation via different explainer LLMs.}
    \label{fig:explainer-models}
\end{figure*}

\begin{figure*}[t]
    \centering
    \includegraphics[width=1\linewidth]{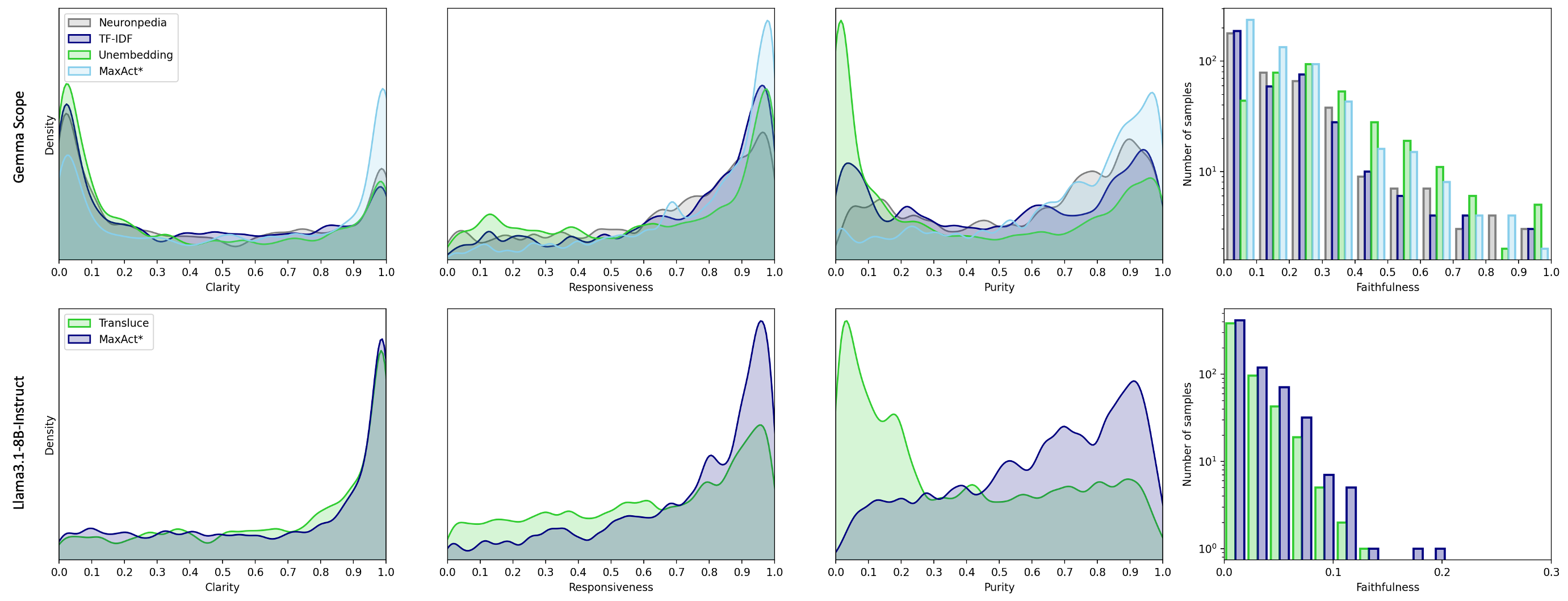}
    \caption{MaxAct* comparison to feature descriptions, generated via other methods on \texttt{Gemma Scope 16K} and \texttt{Llama3.1-8B-Instruct}.}
    \label{fig:baselines}
\end{figure*}

\paragraph{Increasing examples improves performance but raises costs} As shown in Figure~\ref{fig:prompts_analysis} (c), increasing the number of shots consistently enhances performance on activation-based metrics without significantly affecting Faithfulness, contradicting findings in \cite{choi2024automatic}. This discrepancy may stem from differences in provided examples or feature types, as our analysis focuses on \texttt{Gemma Scope SAEs}, whereas \cite{choi2024automatic} examined \texttt{Llama-3.1-8B-Instruct} neurons. Additionally, a higher shot count increases computational costs due to the larger token input (see Table~\ref{tab:token-cost-comparison}). In this study, we primarily use 2-shot prompts, balancing performance and cost efficiency.

\paragraph{Stronger explainer LLMs yield better feature descriptions} More capable models consistently achieve higher performance across nearly all metrics, as presented in Figure \ref{fig:explainer-models}. \texttt{Llama-3.1-70B-Instruct} (quantized) performs comparably to \texttt{GPT-4o mini}, aligning with findings from the evaluation of LLMs (see Table~\ref{tab:evaluation-models-comparison}). Performance generally declines with model size, except for \texttt{Llama-3.2-1B-Instruct}, which frequently fails to adhere to the required output format, leading to significantly poorer results across all metrics.

\paragraph{MaxAct* demonstrates superior performance} Our findings highlight the importance of considering all four \ours\ metrics when optimizing automated interpretability approaches. On \texttt{Gemma Scope 16K}, MaxAct* outperforms all baselines across all metrics except Faithfulness (see Figure~\ref{fig:baselines}). We show that our generated descriptions significantly surpass those currently available on Neuronpedia. Additionally, we compare MaxAct* to TF-IDF and unembedding-based baselines (see Appendix~\ref{appendix-autointerpretability-pipeline}). While the unembedding method underperforms in activation-based metrics such as Clarity, Responsiveness, and Purity, it achieves notably higher Faithfulness by explicitly considering the feature’s output behavior. This underscores that Faithfulness depends on both the feature type (e.g., SAEs vs. MLP neurons) and the generated description.

\begin{table*}[t]
\scriptsize
\centering
\begin{tabular}{p{1.5cm}p{0.45cm}p{0.45cm}p{0.45cm}p{0.45cm}||p{0.45cm}p{0.45cm}p{0.45cm}p{0.45cm}||p{0.45cm}p{0.45cm}p{0.45cm}p{0.45cm}||p{0.45cm}p{0.45cm}p{0.45cm}p{0.45cm}} 
   & \multicolumn{4}{p{3cm}}{MaxAct* 16K} & \multicolumn{4}{p{3cm}}{MaxAct* 65K} & \multicolumn{4}{p{3cm}}{Neuronpedia 16K} & \multicolumn{4}{p{3cm}}{Neuronpedia 65K}   \\ \hline \hline
                & C    & R    & P    & F    & C    & R    & P    & F    & C    & R    & P    & F    & C    & R    & P    & F   \\\hline\hline
Clarity        & 1.00 & 0.68 & 0.63 & 0.17 & 1.00 & 0.69 & 0.65 & 0.14 & 1.00 & 0.63 & 0.53 & 0.21 & 1.00 & 0.56 & 0.49 & 0.28\\ 
Responsiveness  &  -   & 1.00 & 0.86 & 0.16 &  -   & 1.00 & 0.92 & 0.10 &  -   & 1.00 & 0.88 & 0.21 &  -   & 1.00 & 0.87 & 0.14\\
Purity         &  -   &  -   & 1.00 & 0.12 &  -   &  -   & 1.00 & 0.12 &  -   &  -   & 1.00 & 0.24 &  -   &  -   & 1.00 & 0.23\\
Faithfulness   &  -   &  -   &  -   & 1.00 &  -   &  -   &  -   & 1.00 &  -   &  -   &  -   & 1.00 &  -   &  -   &  -   & 1.00 \\ \hline
\end{tabular}
\caption{Correlation of FADE metrics for different sets of feature descriptions.}
\label{tab:metrics-correlation}
\end{table*}

Our results align with \cite{gurarieh2025enhancingautomatedinterpretabilityoutputcentric}, which demonstrates that combining MaxAct-like approaches with output-based methods enhances overall feature description quality. However, the input-centric metric used in that work does not fully capture failure modes that Clarity, Responsiveness, and Purity account for.

This becomes particularly evident when comparing MaxAct* to feature descriptions generated for \texttt{Llama-3.1-8B-Instruct} in \cite{choi2024automatic}. While Clarity scores are comparable -- albeit slightly lower for MaxAct* -- Responsiveness and Purity show significant improvements. This difference may partially stem from the fact, that the dataset used in this work is significantly larger than the one used to produce feature descriptions in \cite{choi2024automatic}. Notably, the Purity distributions of the two approaches are strikingly different, even opposing: MaxAct* exhibits a right-skewed peak, whereas Transluce’s feature descriptions perform poorly on this metric overall. Faithfulness differences are minor but still favor MaxAct*, likely due to the generation of higher-quality feature descriptions.

\paragraph{Metrics correlation depends on the quality of produced feature descriptions} In general, we observe a strong correlation between the evaluated metrics, with the exception of the Faithfulness metric. Importantly, the degree of correlation is influenced by the quality of the feature descriptions being assessed (see Table \ref{tab:metrics-correlation}). Descriptions generated using MaxAct* exhibit the strongest alignment with their corresponding features, whereas Neuronpedia descriptions, especially on \texttt{Gemma Scope 65K}, show the weakest alignment, which results in higher correlation between Clarity, Purity and Responsiveness metrics for MaxAct* Gemma Scope, and lower correlation between Faithfulness and other metrics. 

This trend can be explained by the fact that well-aligned descriptions tend to yield consistently high values across all metrics. Conversely, when the descriptions are of lower quality, the correlation decreases, as each metric is sensitive to different types of failure cases (see Appendix \ref{appendix-qualitative}).

An exception to this pattern is the Faithfulness metric. In its case, the relationship is reversed: poorer description quality leads to a higher correlation with the other metrics. This occurs because Faithfulness is set to zero when other metrics are low and the Faithfulness score is not being computed.

\subsection{Qualitative Analysis}
\label{appendix-qualitative}

\begin{figure}[t]
    \centering
    \includegraphics[width=1\linewidth]{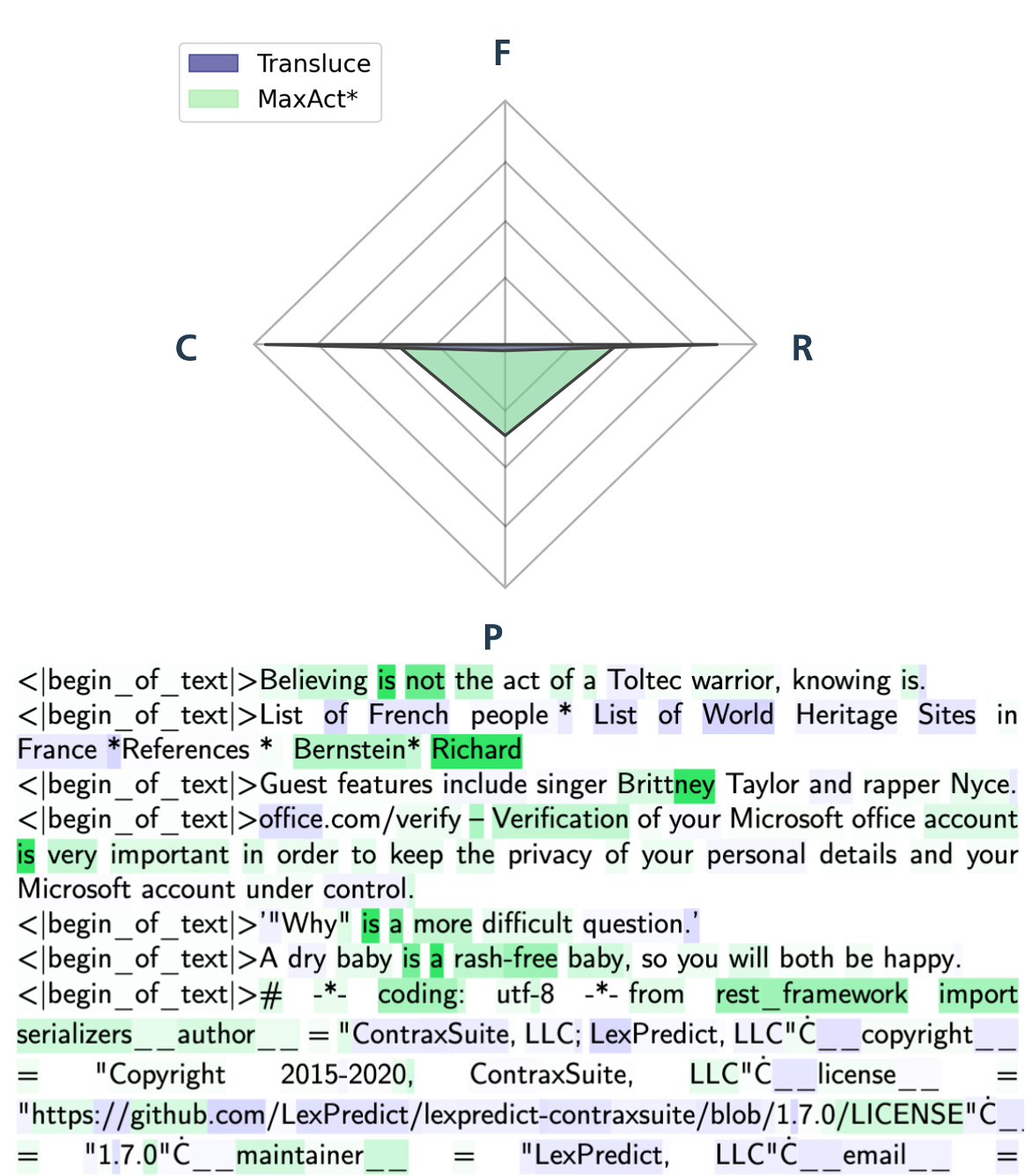}
    \caption{Heatmap and descriptions evaluation result for feature 5183 of \texttt{Llama3.1-8B-Instruct} layer 19.}
    \label{fig:feature-5183}
\end{figure}

\begin{figure}[t]
    \centering
    \includegraphics[width=1\linewidth]{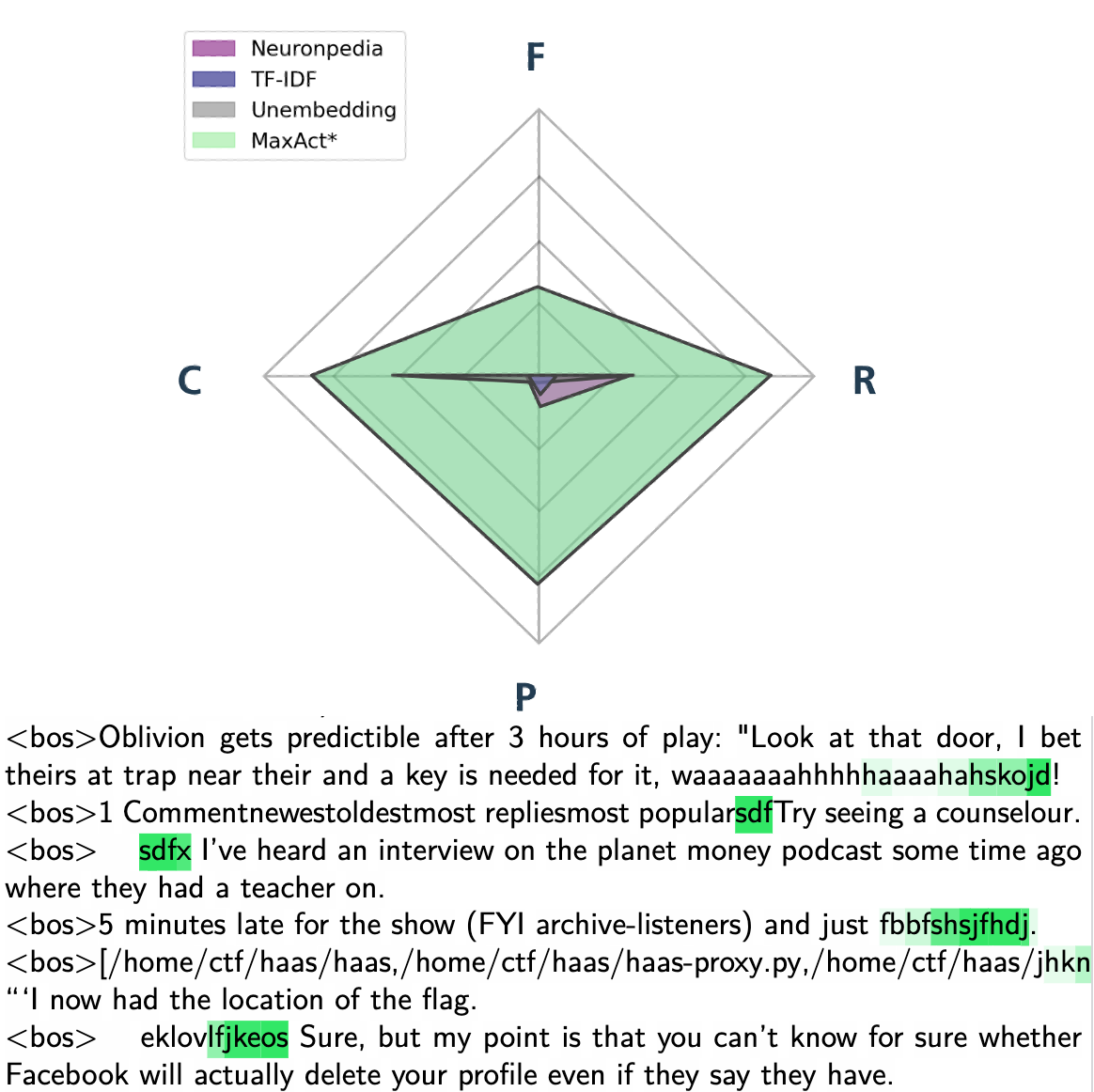}
    \caption{Heatmap and descriptions evaluation result for feature 9295 of \texttt{Gemma Scope} layer 20.}
    \label{fig:feature9295}
\end{figure}

\begin{table}[t]
\scriptsize
\centering
\begin{tabular}{p{1.5cm}p{5.5cm}}
Method & Label  \\ \hline \hline
Transluce & activation on names with specific formatting, including "Bryson," "Brioung," "Bryony," and "Brianna"  \\ 
MaxAct* & Presence and significance of the word "is" \\ \hline
\end{tabular}
\caption{Descriptions for \texttt{Llama3.1-8B-Instruct} feature 5183 layer 19.}
\label{tab:feature-5183}
\end{table}

 \begin{figure*}
    \centering
    \includegraphics[width=1\linewidth]{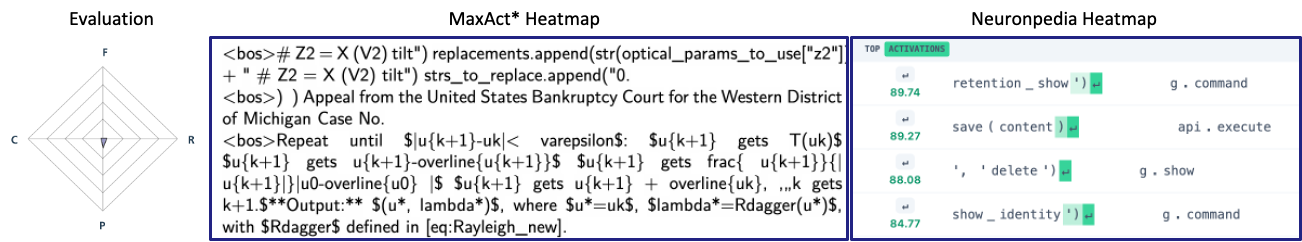}
    \caption{Feature 3,286 of \texttt{Gemma Scope 65K} SAE. MaxAct* description: ``Mathematical expressions and significant numerical values''. Neuronpedia description: ``function definitions in a programming context''. }
    \label{fig:new-line-example}
\end{figure*}

\paragraph{Fine-tuning automated interpretability requires consideration of all metrics in \ours} The comparison of metric distributions against the simulated activation-based metric in Figure \ref{fig:metrics-comparison} highlights that relying solely on this metric is insufficient for accurately assessing the quality of feature descriptions. If an automated interpretability framework is fine-tuned exclusively on such a metric, it may generate suboptimal descriptions.

For instance, evaluations indicate that the description for feature 5183 in layer 19 of \texttt{Llama3.1-8B-Instruct}, as generated in \cite{choi2024automatic}, performs well in terms of Clarity and Responsiveness, yet achieves a near-zero Purity score (see Figure \ref{fig:feature-5183}). Conversely, a description produced using MaxAct* (see Table \ref{tab:feature-5183}) exhibits lower Clarity and Responsiveness but significantly higher Purity.

The heatmap of activations during description generation suggests that the description from \cite{choi2024automatic} strongly activates this feature. While the heatmap does not show all the names listed in Transluce’s description, this may be due to the sampling method, which selects random sentences from the top 1000 to mitigate outliers. However, activation is observed on similar names, such as “Bernstein” and “Brittney.” More importantly, this clearly polysemantic feature responds to multiple distinct concepts, including the word “is” in specific contexts (included into a description generated via MaxAct*), as well as certain coding patterns. 

As a result, despite the relatively high metric score of 0.77 in Transluce’s evaluation, the description has very low Purity. This underscores the importance of considering not only how well a concept activates a feature but also other interpretability factors measurable with \ours. In this case, although the feature is inherently difficult to interpret, we argue that the MaxAct* description provides a more accurate representation, as it better captures the feature’s activating pattern, and \ours\ is clearly demonstrating the feature's polysemanticity.

\begin{table}[h]
\scriptsize
\centering
\begin{tabular}{p{1.5cm}p{5.5cm}}
Method & Label  \\ \hline \hline
Neuronpedia & The presence of JavaScript code segments or functions  \\
TF-IDF & asdfasleilse asdkhadsj easy file jpds just mean span think \\
Unembedding & f,  <eos>, fd, wer, sdf, df, b, jd, hs, ks\\
MaxAct* & Presence of nonsensical or random alphanumeric strings \\ \hline
\end{tabular}
\caption{Descriptions for \texttt{Gemma Scope} feature 9295 layer 20.}
\label{tab:feature9295}
\end{table}

\paragraph{Out-of-Distribution Features in \texttt{Gemma Scope 65K} SAEs} The dataset used for automatic interpretability omitted certain concepts, such as "new line," leading to gaps in feature descriptions. These omissions contribute to the small left-side peak in Responsiveness distribution in Figure~\ref{fig:sae-sizes}. Several features, including 3315, 3858, and 4337, lack activation heatmaps under the MaxAct* approach, as the dataset does not represent their concepts. Consequently, the Explainer Model, relying on unrelated sentences, generates incorrect descriptions (see Figure~\ref{fig:new-line-example}). Heatmaps from Neuronpedia\footnote{https://www.neuronpedia.org/gemma-2-2b/20-gemmascope-res-65k/3286} reveal what would activate these features, highlighting limitations of the dataset used in this work, and broader issues in the automated interpretability pipeline. For example, despite obtaining and visualizing correct results, feature descriptions available on Neuronpedia are also not representing a correct concept. Similar results have been obtained for the <bos> token and indentation in text and code.

\begin{figure}[t]
    \centering
    \includegraphics[width=1\linewidth]{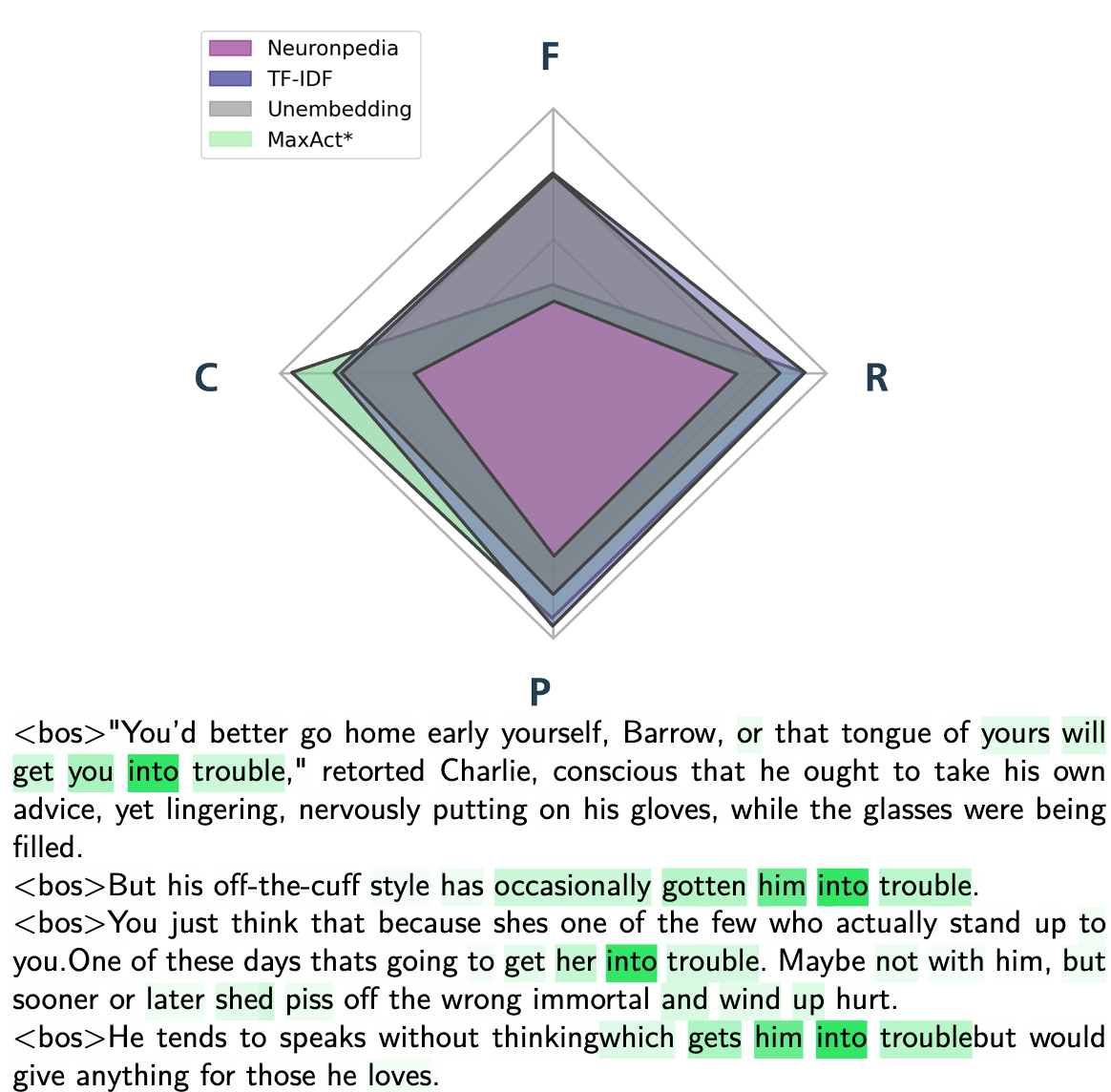}
    \caption{Heatmap and descriptions evaluation result for feature 1139 of \texttt{Gemma Scope} layer 20.}
    \label{fig:feature 1139}
\end{figure}

\begin{table}[t]
\scriptsize
\centering
\begin{tabular}{p{1.5cm}p{5.5cm}}
Method & Label  \\ \hline \hline
Neuronpedia & references to problematic situations or conflicts that cause trouble  \\
TF-IDF & trouble \\
Unembedding & troubles, difficulties, problems, troublesome, mischief, ...\\
MaxAct* & Activation of "into" and "trouble" indicating situations leading to problem \\ \hline
\end{tabular}
\caption{Descriptions for \texttt{Gemma Scope} feature 1139 layer 20.}
\label{tab:feature1139}
\end{table}

\paragraph{Reliable Evaluation -- \ours\ Identifies the Best Description} Different automated interpretability methods prioritize either activation-based metrics or Faithfulness-based measures, leading to descriptions that may be overly broad or inaccurate. 

In some cases, even manual inspection of heatmaps fails to fully capture the underlying concept represented by a feature. Therefore, a comprehensive evaluation must consider all four metrics. 

Table~\ref{tab:feature9295} presents feature descriptions generated by various methods. Based on the heatmap analysis, the MaxAct* description most accurately represents the concept. The unembedding method, while incorporating specific tokens promoted by the feature, also demonstrates strong alignment with the concept, as reflected in the corresponding metrics. However, it is not descriptive enough, which is resulting in lower Responsiveness and Purity.

Sometimes baseline methods may outperform more complex approaches, particularly on specific metrics. For instance, TF-IDF and unembedding baselines exhibit significantly higher Faithfulness compared to Neuronpedia or MaxAct* for certain features (see Figure \ref{fig:feature 1139}).

Feature 1139, for example, influences the output of tokens related to the concept of ``trouble''. Descriptions that explicitly capture this aspect tend to achieve higher Faithfulness (see Table \ref{tab:feature1139}). The MaxAct* description, in contrast, emphasizes the broader meaning and the most activating expression, ``into trouble'', leading to higher Clarity.